\newif\ifhighlightadded
\highlightaddedfalse
\ifhighlightadded
  \DeclareRobustCommand{\addedcolor}{\color[rgb]{0.8,0.1,0.1}}
\else
  \DeclareRobustCommand{\addedcolor}{}
\fi
\DeclareRobustCommand{\added}[1]{{\addedcolor\relax#1}}

\documentclass{assets/matterlab}

\usepackage{microtype}
\usepackage{graphicx}
\usepackage{booktabs}
\usepackage{float}
\usepackage{xurl}
\usepackage{hyperref}
\usepackage{titlesec}

\usepackage{amsmath}
\usepackage{amssymb}
\usepackage{mathtools}
\usepackage{amsthm}
\usepackage{bm}

\usepackage[noabbrev,nameinlink]{cleveref}

\usepackage{orcidlink}
\usepackage{comment}
\usepackage[version=4]{mhchem}
\usepackage{longtable}
\usepackage{array}
\usepackage{multirow}           %
\usepackage{amsfonts}           %
\usepackage{nicefrac}
\usepackage{duckuments}         %
\usepackage{thmtools,thm-restate}
\usepackage{enumitem}
\usepackage{xcolor}
\usepackage{tikz}
\usepackage{tikz-cd}
\usepackage{caption}
\usepackage{subcaption}
\usepackage{listings}
\usepackage{gensymb}
\usepackage{textcomp} %
\usepackage[inkscapelatex=false]{svg}

\newcommand{\addressCHEM}{Department of Chemistry, University of Toronto, Toronto, ON, Canada}
\newcommand{\addressAC}{Acceleration Consortium, Toronto, ON, Canada}
\newcommand{\addressCS}{Department of Computer Science, University of Toronto, Toronto, ON, Canada}
\newcommand{\addressVECTOR}{Vector Institute for Artificial Intelligence, Toronto, ON, Canada}
\newcommand{\addressMSE}{Department of Materials Science \& Engineering, University of Toronto, Toronto, ON, Canada}
\newcommand{\addressCHEMENG}{Department of Chemical Engineering \& Applied Chemistry, University of Toronto, Toronto, ON, Canada}
\newcommand{\addressMEDICALSCI}{Institute of Medical Science, University of Toronto, Toronto, ON, Canada}
\newcommand{\addressCIFAR}{Canadian Institute for Advanced Research (CIFAR), Toronto, ON, Canada}
\newcommand{\addressNVIDIA}{NVIDIA}

\newcommand{\acknowAC}{This research is part of the University of Toronto’s Acceleration Consortium, which receives funding from the CFREF-2022-00042 Canada First Research Excellence Fund. }
\newcommand{\acknowGEN}[1]{This research was enabled in part by support provided by #1 and the Digital Research Alliance of Canada (\url{https://www.alliancecan.ca}). }
\newcommand{\acknowSciNet}[1]{Computations were performed on the #1 supercomputer at the SciNet HPC Consortium. SciNet is funded by: the Canada Foundation for Innovation; the Government of Ontario; Ontario Research Fund - Research Excellence; and the University of Toronto.}

\newcommand{\acknowDARPA}{This work was supported by the Defense Advanced Research Projects Agency (DARPA) under Agreement No.~HR0011262E022.}

\providecommand{\doi}[1]{\href{https://doi.org/#1}{doi:#1}}

\usepackage[utf8]{inputenc}
\usepackage[T1]{fontenc}
\usepackage{url}
\usepackage{booktabs}
\usepackage{amsfonts}
\usepackage{nicefrac}
\usepackage{microtype}
\usepackage{bbm}

\usepackage{amsmath,amsfonts,bm,amssymb}

\def\eqref#1{equation~\ref{#1}}

\def\1{\bm{1}}

\def\vc{{\bm{c}}}

\def\vh{{\bm{h}}}

\def\vl{{\bm{l}}}

\def\vp{{\bm{p}}}
\def\vq{{\bm{q}}}

\def\vu{{\bm{u}}}
\def\vv{{\bm{v}}}

\def\vx{{\bm{x}}}

\def\vz{{\bm{z}}}

\def\mC{{\bm{C}}}

\def\mE{{\bm{E}}}
\def\mF{{\bm{F}}}

\def\mL{{\bm{L}}}

\def\mR{{\bm{R}}}

\def\mPi{{\bm{\Pi}}}

\DeclareMathAlphabet{\mathsfit}{\encodingdefault}{\sfdefault}{m}{sl}
\SetMathAlphabet{\mathsfit}{bold}{\encodingdefault}{\sfdefault}{bx}{n}

\def\sR{{\mathbb{R}}}

\def\sZ{{\mathbb{Z}}}

\newcommand{\E}{\mathbb{E}}
\newcommand{\Ls}{\mathcal{L}}
\newcommand{\R}{\mathbb{R}}

\DeclareMathSymbol{\shortminus}{\mathbin}{AMSa}{"39}

\usepackage{bm}
\usepackage{parskip}

\usepackage{tikz}
\usepackage{amsmath}
\usepackage{makecell}

\newcommand{\hdashline}{%
  \specialrule{0pt}{\aboverulesep}{0pt}%
  \noalign{%
    \hbox to \hsize{%
      \leaders\hbox{%
        \vrule height \lightrulewidth depth 0pt width 2pt%
        \kern 2pt%
      }\hfill
    }%
  }%
  \specialrule{0pt}{0pt}{\belowrulesep}%
}

\usepackage{algorithm}
\usepackage[noend]{algpseudocode}
\usepackage{algorithmicx}

\usepackage{multirow}
\colorlet{algcolor}{mattergreen2}  %
\usepackage{placeins}
\usepackage{graphicx}
\usepackage[percent]{overpic}
\usepackage{tabularx}

\renewcommand{\cite}{\citep}

\usepackage{cleveref}

\newcommand{\name}{\textsc{Clari}}

\usepackage{soul}
\sethlcolor{pink}
\usepackage{etoolbox}

\usepackage{enumitem}
\setlist{leftmargin=0.25in}

\usepackage{siunitx}
\DeclareSIUnit\angstrom{\text{\AA}}

\newcommand{\preprintonly}[1]{#1}
\newcommand{\neuripsonly}[1]{}

\title{Fast Organic Crystal Structure Prediction \\ with Unit Cell Flow Matching}

\author[1,\dagger]{Alston Lo}
\author[2,\dagger]{Luka Mucko}
\author[3,4,5,\dagger]{Austin H. Cheng}
\author[3]{Andy Cai}
\author[3]{Alastair J. A. Price}
\author[1]{Wojciech Matusik}
\author[3,4,5,6,7,8,9,10,11]{Al\'an Aspuru-Guzik}

\affiliation[1]{MIT CSAIL, Cambridge, MA, USA}
\affiliation[2]{University of Zagreb, Faculty of Electrical Engineering and Computing, Zagreb, Croatia}
\affiliation[3]{\addressCHEM}
\affiliation[4]{\addressCS}
\affiliation[5]{\addressVECTOR}
\affiliation[6]{\addressMSE}
\affiliation[7]{\addressCHEMENG}
\affiliation[8]{\addressMEDICALSCI}
\affiliation[9]{\addressAC}
\affiliation[10]{\addressCIFAR}
\affiliation[11]{\addressNVIDIA}

\contribution[\dagger]{Equal contribution}

\metadata[Code]{\url{https://github.com/the-matter-lab/clari}}

\abstract{Organic crystal structure prediction (CSP) is a requirement for computational modelling of organic solids, but traditionally costs several CPU-years per molecule.
Generative models such as OXtal dramatically reduce this cost by sampling stable organic crystal structures directly.
However, OXtal forgoes explicit lattice parametrization in favour of modelling large crops of the bulk material with expensive triangle layers, which can incur a computational cost of minutes per molecule.
In this paper, we reduce this to seconds with \name{}, a large-scale flow matching model that generates redundancy-free unit cells and replaces triangle layers with pair-bias attention.
\name{} requires only atom types and bonds as input and does not need an RDKit-sanitizable input molecule, which expands its applicability to challenging chemistries such as fullerenes, metal complexes, and atom clusters.
We further ablate key design choices such as auxiliary losses, timestep distributions, noise priors, and self-conditioning.
On OXtal's test sets, we surpass OXtal's solve rate while obtaining a speedup of $15$--$30\times$.
Because \name{} also models explicit hydrogens, it supports inference-time scaling via direct energy ranking, without any decoration or relaxation step.
When generating 150 crystals and selecting the top-30 by energy, we further improve solve rate while maintaining a speedup of $5$--$8\times$.
We also introduce the CSD Teaching Subset as a new test split of diverse and complex molecules for future benchmarking.
Our contributions enable CSP within seconds, making large-scale virtual screening of organic solids practical.
}

\begin{document}

\maketitle

\section{Introduction}

The properties of organic solids depend strongly on crystal packing.
Knowing a molecule's crystal structure is therefore a prerequisite for modelling solid-state properties in applications spanning fertilizers \cite{honer2017mechanosynthesis}, pesticides \cite{yang2017ddt}, pigments \cite{hao1997some, panina2008polymorph}, food \cite{aguilera2008food}, energetic materials \cite{arnold2023crystal}, pharmaceuticals \cite{price2016can, bowskill2021crystal}, and organic electronics \cite{forrest2004path} such as organic light-emitting diodes \cite{sun2023efficient} and organic photovoltaics, as well as emerging flexible materials \cite{bhattacharya2023atomistic,koshima2011photomechanical}.
Crystal structures also enable the design of templates that seed desired crystal forms of other molecules \cite{chadwick2011polymorphic,buvcar2013curious}.
\added{Yet in many practical settings, no experimental crystal structure is available, so it must be predicted computationally.}
\preprintonly{\pagebreak}

The problem of crystal structure prediction (CSP) has challenged scientists for decades \cite{maddox1988crystals}, owing to the large and rugged search space of molecular crystals.
Traditional CSP pipelines tackle this by executing a dense computational funnel.
Starting from a large set of randomly generated candidate structures, they iteratively filter and re-rank candidates using successively more accurate but costly quantum chemistry methods, culminating in density functional theory (DFT) relaxations \cite{hunnisett2024seventh_gen, hunnisett2024seventh_rank, BeranCrystals}.
While these methods are accurate \cite{hoja2019reliable}, exhaustive search pipelines can require several CPU-years per molecule \cite{zhou2025robust, reilly2016report}, making them prohibitive for screening even modestly sized virtual libraries.
In practice, this bottleneck confines traditional CSP to targeted pharmaceutical campaigns \cite{mortazavi2019computational} and renders large-scale screening on the basis of solid-state properties all but infeasible \cite{ishii2020charge}.

A given molecule can form multiple stable crystal forms in a phenomenon known as \emph{polymorphism} \cite{bernstein2020polymorphism}.
The one-to-many relationship between chemical identity and crystal polymorphism makes the problem map closely to generative modelling.
Indeed, recent work has introduced generative models as a direct alternative to energy-based search.
OXtal \cite{jin2025oxtal} demonstrated that a generative model can learn to sample the distribution of experimentally observed organic crystal structures end-to-end.
While paradigm-shifting, OXtal inherits a number of expensive design choices from AlphaFold3 \cite{abramson2024accurate}: it represents crystals in bulk form, processing multiple symmetry-related copies of each molecule simultaneously, and it relies on triangle-update layers \cite{jumper2021highly} to propagate pairwise geometric information.
Together, these choices impose a large computational footprint that makes training and inference costly, and poses barriers to screening ultra-large libraries.

\begin{figure}[t]
\centering
\centerline{\includegraphics[width=\textwidth]{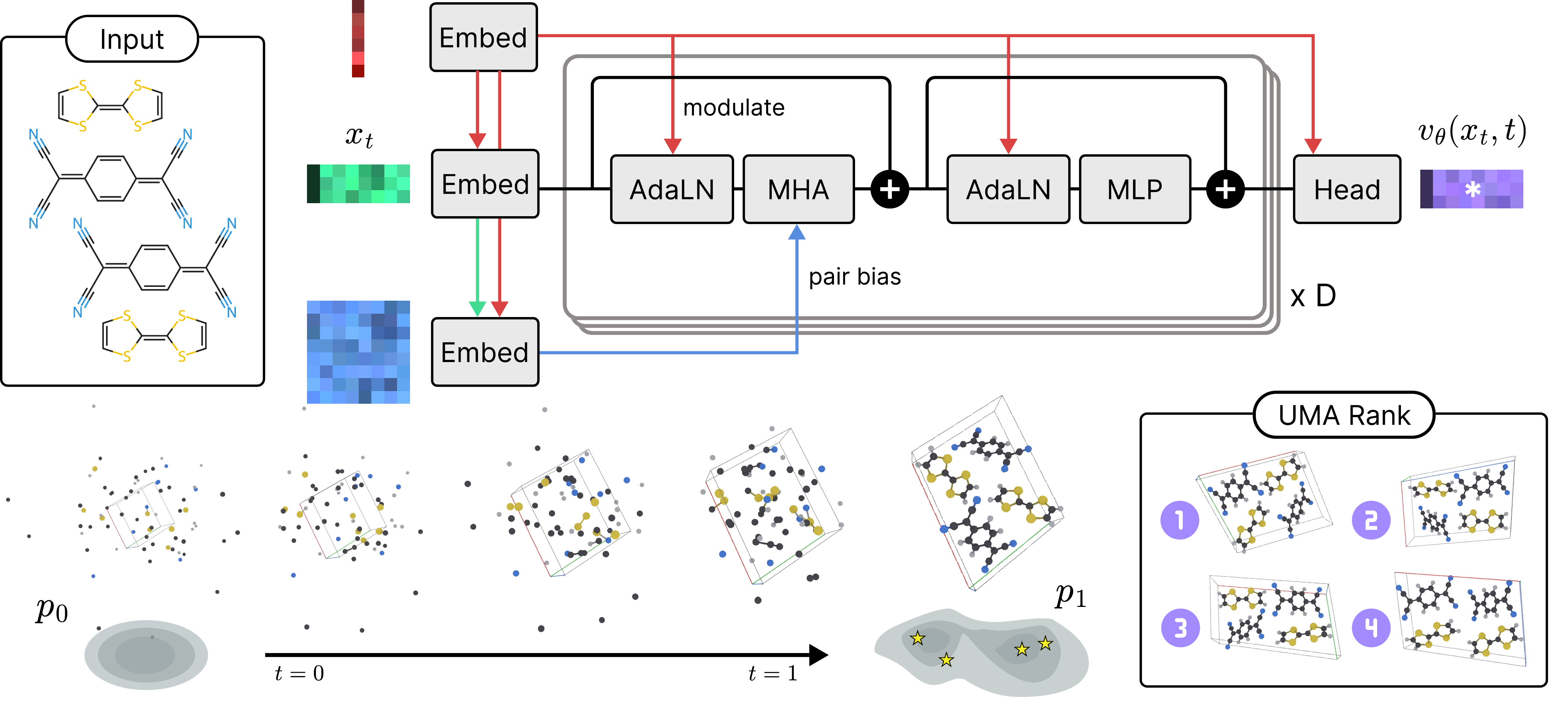}}
\caption{Overview of \name{}. Given a molecular graph as input, \name{} directly generates a single unit cell comprising atom coordinates and lattice vectors via a flow matching trajectory. No bulk expansion or triangle layers are required.}
\label{fig:schematic}
\vspace{-0.1in}
\end{figure}

In this work, we show that two targeted refinements suffice to \added{narrow} the gap between generative CSP and practical screening.
First, rather than modelling a bulk crop of symmetry-equivalent copies, we train \name{} directly on \emph{unit cells}, jointly predicting atom coordinates and lattice vectors under a flow matching objective\added{ (Figure~\ref{fig:schematic})}.
Second, we \emph{replace triangle-update layers} with a pair-bias attention architecture, which transmits pairwise geometric information through attention logits without the cubic cost of triangle operations.
\added{Forgoing bulk crops and triangle layers may make global consistency harder to learn, but the resulting gain in efficiency enables more training updates and faster inference.}
Together, these choices \added{allow} \name{} \added{to} surpass OXtal in prediction quality while also accelerating sampling by roughly $15$--$30\times$ on the CSP Blind Tests, or $5$--$8\times$ end-to-end when ranking by energy with the Universal Model for Atoms (UMA) \cite{wood2025family}.
We also train on all-atom structures including hydrogens, which enables direct energy-based ranking without any decoration or relaxation step.
Ablations isolate the contribution of each component: self-conditioning, the choice of lattice source distribution, \added{auxiliary losses}, and the timestep schedule.

\begin{figure}[t]
\centering
\newcommand{\rowheight}{4cm}
\begin{minipage}[c][\rowheight][c]{0.24\textwidth}
  \centering
  \includegraphics[width=0.95\linewidth]{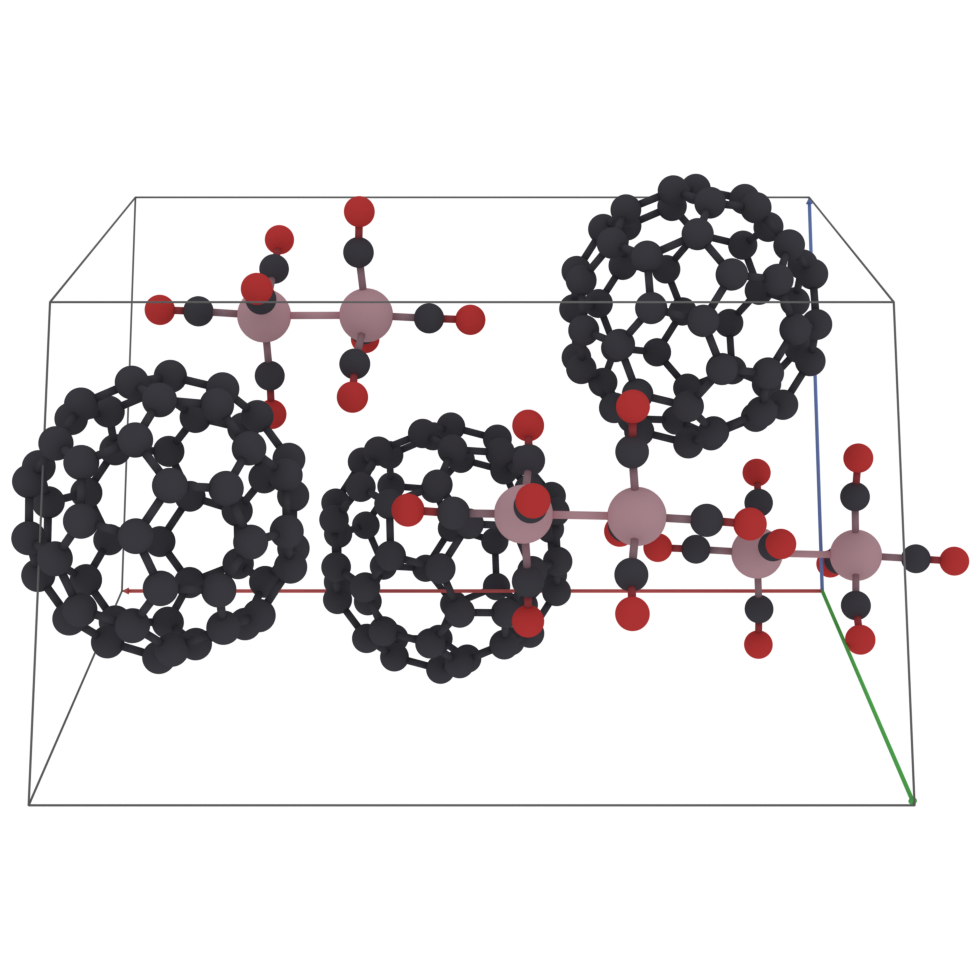}
\end{minipage}\hfill
\begin{minipage}[c][\rowheight][c]{0.24\textwidth}
  \centering
  \includegraphics[width=0.8\linewidth]{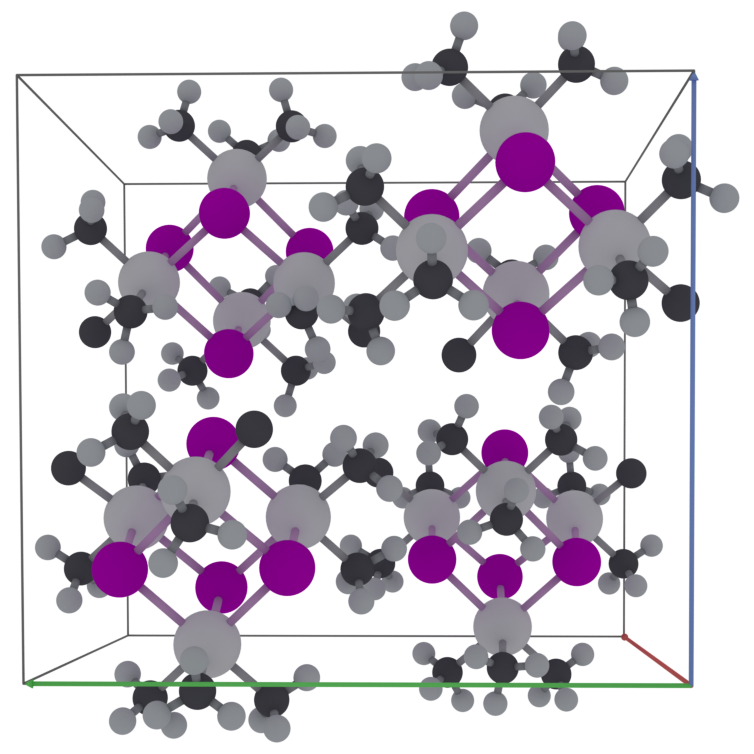}
\end{minipage} \hfill
\begin{minipage}[c][\rowheight][c]{0.24\textwidth}
  \centering
  \includegraphics[width=0.9\linewidth]{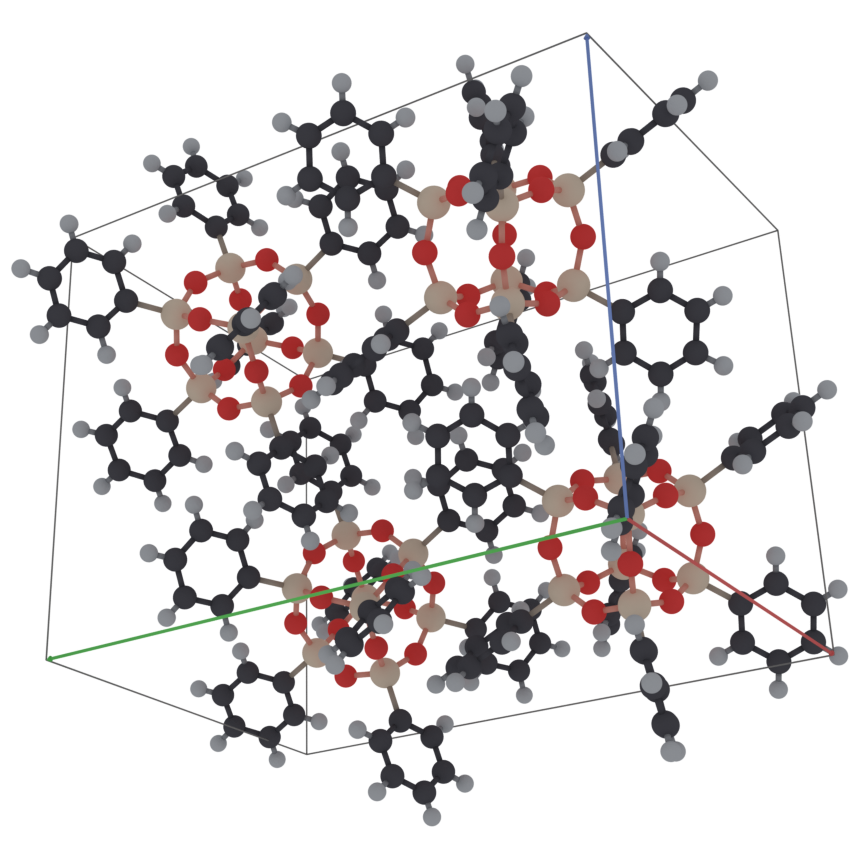}
\end{minipage}\hfill
\begin{minipage}[c][\rowheight][c]{0.24\textwidth}
  \centering
  \includegraphics[width=0.95\linewidth]{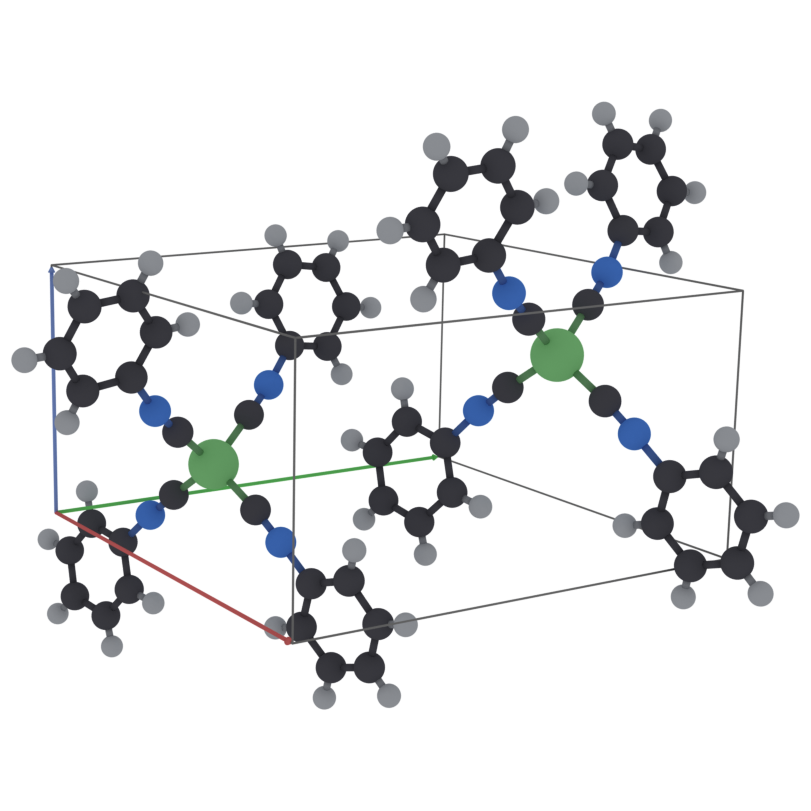}
\end{minipage}

\begin{minipage}{0.24\textwidth}\centering\small (a) YUGWUT\end{minipage}\hfill
\begin{minipage}{0.24\textwidth}\centering\small (b) VISMOB\end{minipage}\hfill
\begin{minipage}{0.24\textwidth}\centering\small (c) DORRAF\end{minipage}
\hfill
\begin{minipage}{0.24\textwidth}\centering\small (d) XADQUP\end{minipage}
\vspace{0.05in}

\caption{Crystal structures predicted by \name{} across chemically diverse classes: (a) YUGWUT, a $\mathrm{C}_{60}\!\cdot\!\mathrm{Co}_2(\mathrm{CO})_8$ fullerene cocrystal; (b) VISMOB, the $[(\mathrm{CH}_3)_3\mathrm{PtI}]_4$ trimethylplatinum iodide tetramer with a $\mathrm{Pt}_4(\mu_3\text{-}\mathrm{I})_4$ cubane core; (c) DORRAF, an octaphenyl-substituted $\mathrm{Si}_8\mathrm{O}_{12}$ POSS cage; (d) XADQUP, a tetrahedral transition-metal complex.}
\label{fig:hero}
\vspace{-0.1in}
\end{figure}

In summary, our main contributions are:
\begin{itemize}
  \item We introduce \name{}, a flow matching model for organic CSP that operates on a single unit cell with a pair-bias attention architecture, sampling roughly $15$--$30\times$ faster than OXtal on the CSP Blind Tests ($5$--$8\times$ end-to-end with UMA energy ranking). Inference takes on the order of seconds per molecule. \added{Unlike OXtal, \name{} does not require input conformers, which can be cumbersome to obtain for complex chemistries like metal complexes and atom clusters (Figure~\ref{fig:hero}).}
  \item The resulting speed opens generative CSP to virtual library screening at scale \cite{omar2021high}, a regime previously accessible only to fast but coarse traditional heuristics. This throughput also facilitates inference-time scaling, including energy-based ranking via joint heavy-atom and hydrogen modelling.
  \item For a rigorous final evaluation, we combine complementary metrics into a single evaluation suite and introduce a new test split drawn from the CSD Teaching Subset, covering chemically complex systems including fullerenes, boranes, and organometallic complexes that were excluded from or underrepresented in prior benchmarks. We also conduct systematic ablations of architectural and training choices, revealing which components drive performance gains.
\end{itemize}

\section{Related work}

\textbf{Computational (simulation-based) organic crystal structure prediction.}
Organic CSP has historically been framed as a two-stage pipeline consisting of structure generation followed by energy ranking. This paradigm has been formalized and benchmarked through the Cambridge Crystallographic Data Centre (CCDC) Blind Tests \cite{lommerse2000test, motherwell2002crystal, day2005third, day2009significant, bardwell2011towards, reilly2016report, hunnisett2024seventh_gen,hunnisett2024seventh_rank}, which evaluate CSP methods on unpublished experimental structures. 
Early approaches relied on extensive exploration of the configurational space using quasirandom sampling \cite{lin2016structure,case2016convergence}, simulated annealing \cite{catlow1993simulating}, and evolutionary algorithms \cite{curtis2018evolutionary}. While these methods are general, they require generating and evaluating a large number of candidate structures, making CSP computationally expensive in practice.

\textbf{Machine learning for accelerating CSP.}
Subsequent work has focused on reducing the cost of the ranking stage rather than altering the search paradigm. In particular, machine learning interatomic potentials (MLIPs) have been widely adopted as surrogates for DFT, significantly accelerating energy and force evaluations \cite{hunnisett2024seventh_rank}. Systems such as FastCSP \cite{gharakhanyan2025fastcsp} demonstrate that MLIPs can speed up traditional pipelines by orders of magnitude. However, these approaches still filter large candidate sets, leaving the combinatorial burden of structure generation largely unchanged.

\textbf{Generative CSP.}
More recently, generative approaches have emerged as an alternative to exhaustive search, aiming to directly sample low-energy crystal structures. OXtal \cite{jin2025oxtal} introduced this paradigm for organic CSP using an AlphaFold3-inspired architecture. Follow-up work explores different generative formulations, including reinforcement learning in PackFlow \cite{subramanian2026packflow} and flow matching on rigid bodies in MolCrystalFlow \cite{zeng2026molcrystalflow}. These methods demonstrate that sufficiently expressive generative models can substantially reduce the reliance on downstream ranking. However, current approaches either employ computationally intensive architectures (e.g., triangle updates over large crops) or rely on restrictive representations such as rigid molecules, limiting efficiency and applicability.

\textbf{Generative models for chemistry.} Outside of organic CSP,
parallel work in inorganic materials has developed a range of generative models for periodic crystals, beginning with variational approaches such as CDVAE \cite{xie2022crystal} and followed by models that more explicitly incorporate symmetry and periodicity \cite{jiao2023crystal,jiao2024space,miller2024flowmm,zeni2025generative,hollmer2025open,luo2025crystalflow,levy2025symmcd}.
A consistent design choice across these methods is the use of \emph{explicit lattice \added{parametrizations}}, in which the unit cell and atomic positions are modelled directly. Our work is also related to advances in generative modelling of molecular geometry, including proteins \cite{watson2023novo, yim2023se,bose2024sestochastic,jing2023eigenfold,jing2024alphafold, geffner2025proteina, geffner2025laproteina}, biomolecular complexes \cite{abramson2024accurate, didi2026scaling}, and molecules \cite{stark2024harmonic,wang2024swallowing,dunn2026flowmol3,irwin2025semlaflow,reidenbach2026applications,vonessen2025tabasco}.
These works demonstrate the effectiveness of generative models for structured 3D data, from which we borrow several techniques. \added{LLM-based approaches such as CrystaLLM \cite{antunes2024crystal} instead generate CIF files directly. This representation is flexible and expressive (e.g., for space groups and disorder), but can be considerably less compact than unit cell point clouds, which may limit scalability and efficiency.}

\section{\name{}}
\label{sec:method}

\subsection{Crystal structure prediction}
\label{sec:crystal}

We represent a crystal unit cell as a tuple $(\mL, \mC, \mF, \mE)$, where $\mL  \in \sR^{3 \times 3}$ is a lattice matrix with primitive vectors stored as rows, $\mC \in \sR^{N \times 3}$ are Cartesian atom coordinates with zeroed centroid, $\mF \in \sR^{N \times d}$ are atom features such as atomic numbers, and $\mE \in \sR^{N \times N}$ is an adjacency matrix of bond types. Unit cells periodically tile the space to produce full crystals through translations by $\mL^\top\vz \in \R^3$, for any $\vz \in \sZ^3$. This defines a notion of \textit{periodic} distance $d_\mL(\vp, \vq) = \min_{\vz \in \sZ^3}||\vp - \vq - \mL^\top\vz||_2$, which is the minimum distance between the periodic images of points $\vp$ and $\vq$ in the unit cell.

The problem of crystal structure prediction is to infer the positions $(\mL, \mC)$ given the crystal molecular graph $(\mF, \mE)$.
Concretely, the input is the 2D molecular graph of \added{the} molecules in the unit cell\added{ with multiplicity}; we do not consider the space group as input.
We frame CSP as a conditional generative modelling problem and adopt a flow matching approach due to its simplicity and recent success in biomolecular modelling \cite{li2026flowmatchingmeetsbiology}. A key design choice is that we generate $(\mL, \mC)$ as a unified object by treating the primitive lattice vectors as three additional \textit{virtual} points. That is, we concatenate them row-wise into a matrix $\vx = (\frac{1}{2}\mL; \mC) /\sigma \in \sR^{(3+\added{N}) \times 3}$ with $\sigma$ chosen to normalize $\vx$ to roughly unit variance across the dataset. Assuming $(\mF, \mE)$ is fixed, $\vx$ is invariant under (i) signed permutations of the lattice rows, (ii) permutations of atomic rows that are also automorphisms of the crystal molecular graph, (iii) rotations, and (iv) independent periodic translations of any connected molecular subgraph, which we call a \textit{body} or \textit{component}.

\subsection{Flow matching}
\label{sec:flow}

Flow matching \cite{lipman2022flow, liu2022flow, albergo2022building} learns a continuous-time vector field that transports a tractable source distribution $p_0$ into the data distribution $p_1$.
Adopting the linear interpolant $\vx_t = (1-t)\vx_0 + t\vx_1$, a network $v_\theta$ is trained to regress the target velocity:
\begin{equation}
\Ls_{\mathrm{FM}} = \E_{t,\, \vx_0,\, \vx_1}\!\left[\,\bigl\| v_\theta(\vx_t, t) - (\vx_1 - \vx_0) \bigr\|^2\,\right], \quad t \sim p(t),\; \vx_0 \sim p_0(\vx),\; \vx_1 \sim p_1(\vx).
\end{equation}
In practice, we split $\Ls_{\mathrm{FM}}$ into lattice and coordinate terms that are summed $\Ls_{\mathrm{FM}}^\mL+ \Ls_{\mathrm{FM}}^\mC$. Each term is computed using a mean-squared error so that they contribute equal weight to the final loss. A trained flow matching model $v_\theta$ can then be used to generate samples by evolving $\vx_0 \sim p_0(\vx)$ from the source distribution according to $\mathrm{d}\vx_t = v_\theta(\vx_t, t)\mathrm{d}t$ using any ordinary differential equation (ODE) solver.

\begin{figure}[t]
\centering
\includegraphics[width=\textwidth]{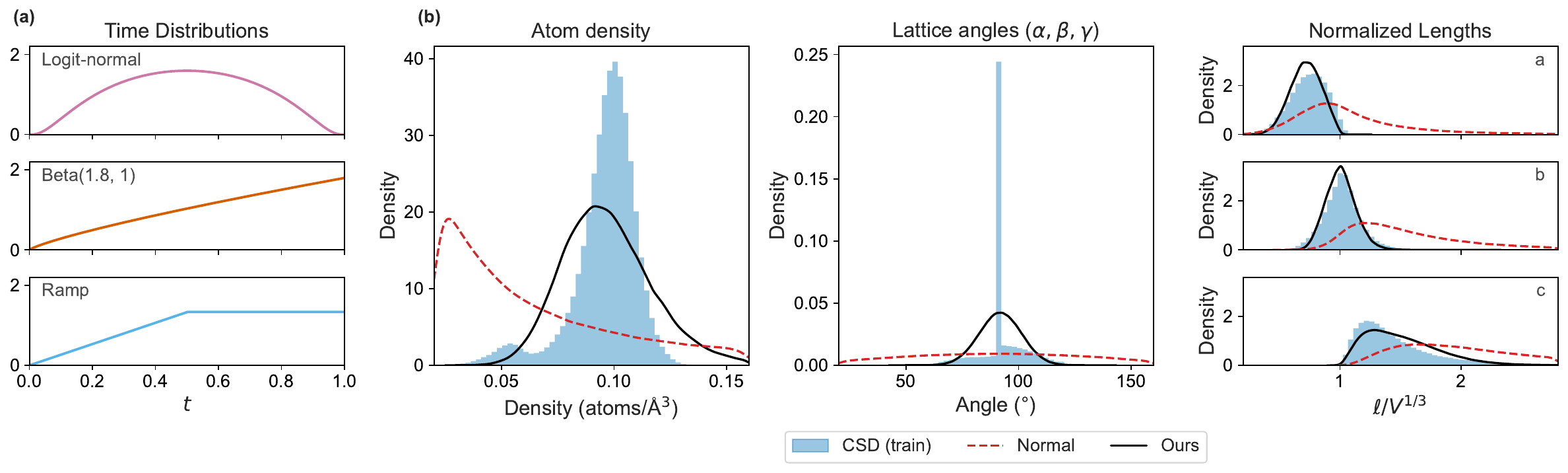}
\caption{\textbf{(a)} Time distributions $p(t)$ considered in our ablations: uniform (not depicted), logit-normal, $\text{Beta}(1.8, 1)$, and ramp. \textbf{(b)} Marginals of the lattice source distribution $p_0$ versus the Cambridge Structural Database (CSD) training set: atom density (atoms/\si{\angstrom}$^3$), the three unit cell angles, and the sorted volume-normalized lattice lengths $\ell / V^{1/3}$. A standard Gaussian source concentrates mass at vanishing density, producing degenerate lattices, and badly misfits the length distribution. Our fitted prior closely tracks the data marginals.}
\label{fig:pt_source}
\end{figure}

\textbf{Time distribution and schedule.}
A rich and important design space lies in choosing the timestep training distribution $p(t)$ and sampling discretization $(t_i)_i$. Prior work has found that skewing towards later timesteps leads to better results in biomolecular modelling \cite{geffner2025proteina}, where fine-grained local details matter more than, for example, in images. However, we argue that crystals also require attention to the overall global arrangement. In Section~\ref{sec:ablations}, we explore a variety of options including uniform, logit-normal \cite{esser2024scaling}, and beta \cite{geffner2025proteina, vonessen2025tabasco} distributions and linear and log discretizations (Figure~\ref{fig:pt_source}a). 

\textbf{Source distribution.} The de facto choice of $p_0$ is a unit normal distribution. However, a random normal matrix with high probability yields unrealistic cell densities and degenerate periodic images, which we find leads to a considerable degradation in performance. Instead, we use a data-informed prior $p_0$ by decomposing the lattice into three components (atom density, cell angles, and volume-normalized cell lengths) and fitting a Gaussian distribution to each using the training data (Figure~\ref{fig:pt_source}b). To sample from $p_0$, we sample each component independently, reconstruct the lattice matrix, and then apply a random rotation and signed permutation. Further details are given in Appendix~\ref{app:model:source}.     

\textbf{Auxiliary losses.}
In order to improve physical quality, we enrich the base flow matching objective with auxiliary losses between the one-step estimate $\hat{\vx}_1 = \vx_t + (1-t)\, v_\theta(\vx_t, t)$ and the ground truth $\vx_1$. In protein models, it is common to use a bond loss, smooth local distance difference test (LDDT), or distogram loss \cite{abramson2024accurate, geffner2025proteina}. We propose a new pair of losses relevant for crystal generation: (i) $\Ls_{\mathrm{vol}}$, the relative error between predicted and target lattice volumes; and (ii) $\Ls_{\mathrm{pair}}$, a pairwise periodic distance error that also penalizes steric clashes. 
The full objective is then $\Ls_{\name{}} = \Ls_{\mathrm{FM}}^{\mL} + \Ls_{\mathrm{FM}}^{\mC} + \Ls_{\mathrm{vol}} + \Ls_{\mathrm{pair}}$. The formal definitions of the auxiliary losses are provided in Appendix~\ref{app:model:losses}.

\textbf{Self-conditioning.} \name{} also uses self-conditioning \cite{chen2023analog, stark2024harmonic, dunn2026flowmol3} whereby the model recycles its own endpoint estimate $\hat{\vx}_1$ from the previous timestep as input.
This incurs effectively no cost during inference, but slows training by $\sim$20\% due to an extra forward pass (on a half batch) every step. Despite this, we find self-conditioning to yield a consistent performance improvement\added{,} so we include it in our final models. 

\subsection{Architecture}
\label{sec:architecture}

An emerging paradigm in biomolecular structure modelling is using large scalable Transformers that are not inherently equivariant but instead learn it through large-scale training with data augmentation \cite{wang2024swallowing, abramson2024accurate,geffner2025proteina}. In the same spirit, \name{} implements $v_\theta$ using the Diffusion Transformer (DiT) \cite{peebles2023scalable} architecture with modern Transformer \cite{vaswani2017attention} tricks such as gated attention \cite{qiu2026gated}, QKNorm \cite{chowdhery2023palm}, and SwiGLU \cite{shazeer2020glu}. Since the lattice is treated as virtual points, the lattice and atom tokens participate identically throughout the network and no special pooling layers are required. Pair features are created from bond, topological and geometric distance information, which modulate the DiT blocks through additive pair attention biasing. Finally, AdaLN-Zero blocks \cite{peebles2023scalable} are used to modulate the DiT with global-level information such as the timestep, lattice, and molecular formulae. Importantly, \name{} contains no triangular operations\added{,} which results in a significantly more scalable and efficient architecture. This allows \name{} to scale to \added{350M} parameters even for unit cells with up to $512$ atoms. Further details are in Appendix~\ref{app:model:architecture}.

\subsection{Optimal transport and augmentation}
\label{sec:augment}

Standard flow matching draws $\vx_0$ and $\vx_1$ independently, but coupling them via an approximate optimal-transport map straightens trajectories and reduces variance of the regression target \cite{tong2023improving, klein2023equivariant, wohlwend2025boltz}. We aim to align $\vx_0$ and $\vx_1$ with respect to (i) signed lattice permutation, (ii) atomic permutation, and (iii) rotations, as described in Section~\ref{sec:crystal}. However, an exact coupling under the joint action is intractable. We therefore approximately align $\vx_1$ to $\vx_0$ by composing three tractable steps that each minimize squared deviation along individual symmetry groups: brute-force signed-permutation lattice alignment, atom-permutation alignment via the Hungarian algorithm \cite{kuhn1955hungarian}, and a final weighted Kabsch \cite{kabsch1976solution} pose alignment. We refer readers to Appendix~\ref{app:model:ot} for further details. Note that alignment effectively performs augmentation due to the invariance of our prior with respect to these actions. We handle the last remaining symmetry of periodic translation (iv) through an augmentation step before alignment. Specifically, we randomly translate the unit cell, re-wrap the centroid of each component to lie in $[0, 1)^3$, and finally re-centre the centroid of the atomic coordinates.

\subsection{Inference-time scaling} \label{sec:ttscale}

For an efficient model like \name{}, \added{a simple way to improve performance is inference-time scaling, also known as} test-time compute \cite{ma2025inference}. Recent work \cite{didi2026scaling} has shown the value of
these techniques for flow-based atomistic generation, applying methods like beam search and Feynman-Kac steering \cite{singhal2025a,skreta2025feynmankac} to binder design. We leave a principled exploration of such methods for future work and, for now, opt for a simple best-of-$N$ approach. Specifically, we generate multiple candidate structures per target, score them with the UMA model \texttt{uma-s-1p2} \cite{wood2025family}, and report results on a subset of lowest-energy structures. Because \name{} generates full all-atom crystals, we can immediately rank candidates by energy without having to perform any hydrogen decoration or relaxation steps.

\section{Experiments}
\label{sec:experiments}

We evaluate \name{} on held-out crystals (\emph{Rigid} and \emph{Flexible} OXtal subsets), three crystal structure prediction (CSP) Blind Tests (CSP5--7), and the Cambridge Structural Database (CSD) Teaching Subset. \added{We begin by comparing \name{} against baselines in Section~\ref{sec:comparison}. Then, we motivate our design choices through a series of ablations that build up to \name{} in Section~\ref{sec:ablations}.}

\subsection{Experimental setup}
\label{sec:experimental-setup}

\textbf{Dataset.}
We train \name{} on the Cambridge Structural Database (CSD) \cite{csd}, a million-scale database of experimentally determined organic and metal-organic crystal structures. \added{We keep entries that have a 3D structure, are non-polymeric, and are derived from} single-crystal diffraction at ambient pressure. For non-test splits, we further filter for entries deposited up to May 1, 2025 (following \citet{jin2025oxtal}), with R-factor below $10$\%, and at most $512$ atoms in the unit cell. 
Unlike prior efforts~\cite{jin2025oxtal, subramanian2026packflow}, we do not remove hydrogens or require RDKit sanitization, allowing us to retain a much larger set of entries. After metadata filtering, we found the dataset to still be relatively noisy due to unflagged polymers or superimposed molecules from improperly specified disorder. We attempt to catch these cases using distance-based thresholds, which are described in Appendix~\ref{app:dataset} among other details.
The OXtal test crystal families and the CSD Teaching Subset are combined to create our held-out test pool.

\textbf{Training and inference.}
We train a medium model \name{}-M ($88$M parameters) for 150,000 steps on $4$ NVIDIA H100 GPUs. We then scale the model to \name{}-L ($173$M parameters) on $8$ H100 GPUs\added{ and further to \name{}-H ($350$M parameters) on $16$ H100 GPUs} for final comparisons against baselines. Hyperparameters are given in Appendix~\ref{app:model:hparams}. During training, refcode families are sampled uniformly to avoid bias towards molecules with many polymorphs. We take the representative with the lowest R-factor from each refcode family for validation and testing. For inference, we use the Heun sampler \cite{karras2022elucidating} with \added{20} and \added{50} steps in Sections \added{\ref{sec:comparison}} and \added{\ref{sec:ablations}}, respectively.

\textbf{Metrics.}
Robustly assessing predicted crystals is difficult \cite{mayo2022development}, so we report a battery of complementary metrics.
\added{For baseline comparisons, we follow the OXtal evaluation protocol with COMPACK \cite{chisholm2005compack}, which searches for matching clusters of molecules between the predicted and ground-truth crystals. Our primary metric is approximate solve rate \textbf{Sol@$\bm{k}$} ($\uparrow$): a target counts as solved if at least one of its $k$ samples matches the ground truth in at least $8$ of $15$ molecules with $\mathrm{RMSD}_{15}<2$\,\AA{} and no inter-body clashes, where $\mathrm{RMSD}_{15}$ is the root-mean-square deviation across the matched 15-molecule cluster. Throughout the paper, Sol refers to this $\geq 8/15$ variant, matching OXtal for direct comparison. The strict $15/15$ variant is reported in Table~\ref{tab:sol-strict-all}.}
For ablations, we use \added{cheaper metrics}:
(1) \textbf{Clash Rate} (\%, $\downarrow$), the fraction of structures with an inter-body atom pair closer than their sum of covalent radii;
(2) \textbf{PoseBusters} pass rate (\%, $\uparrow$) \cite{buttenschoen2024posebusters} on applicable fragments;
(3) relative \textbf{Volume Error} (\%, $\downarrow$); and
(4) \textbf{EMD PDD} ($\downarrow$), an earth mover's distance (EMD) between pointwise distance distributions (PDD) \cite{widdowson2022resolving}, an isometry invariant of periodic point sets.
We split these into \emph{quality} (Clash Rate, PoseBusters) and \emph{reconstruction} (Vol. Error, EMD PDD) metrics.

\newcommand{\dftavg}{DFT$_{\mathrm{avg}}$}

\begin{table}[t]
\centering
\small
\caption{\added{Approximate solve rate} Sol@$k$ across the OXtal benchmark subsets. $n_s$ is the generation budget (samples drawn from the model), while $k$ is the number used to compute Sol@$k$. \dftavg{} is the average across participants in the corresponding CSP Blind Test as evaluated by \citet{jin2025oxtal} (they use $n_s = k = 464 / 83/ 868$ for CSP5/CSP6/CSP7, respectively). When $k<n_s$, our methods select the top-$k$ by UMA energy. Best per column \added{within each block of our rows} in \textbf{bold}. Bootstrap
means over 5000 resamples are reported, and standard errors are given in Table~\ref{tab:sol-base-all}.}
\label{tab:all-metrics}
\begin{tabular}{lrr c c c c c c}
\toprule
 & & & Rigid & Flexible & CSP5 & CSP6 & CSP7 & Teach. \\
Method & $n_s$ & $k$ & (50) & (50) & (6) & (5) & (8) & (773) \\
\midrule
\dftavg{} & -- & -- & -- & -- & 0.544 & 0.496 & 0.421 & -- \\
\midrule
OXtal & 30 & 30 & 0.300 & 0.220 & 0.167 & 0.200 & 0.125 & -- \\
\name{}-M & 30 & 30 & 0.697 & 0.241 & 0.554 & 0.311 & 0.210 & 0.442 \\
\name{}-L & 30 & 30 & 0.731 & 0.287 & 0.681 & 0.355 & 0.245 & 0.461 \\
\name{}-H & 30 & 30 & 0.766 & 0.378 & 0.705 & 0.445 & 0.280 & 0.584 \\
\name{}-M & 150 & 30 & 0.712 & 0.260 & 0.616 & 0.374 & 0.225 & 0.470 \\
\name{}-L & 150 & 30 & 0.772 & 0.346 & \textbf{0.789} & 0.480 & 0.263 & 0.484 \\
\name{}-H & 150 & 30 & \textbf{0.798} & \textbf{0.479} & 0.785 & \textbf{0.612} & \textbf{0.393} & \textbf{0.621} \\
\midrule
\name{}-M & 400 & 200 & 0.879 & 0.506 & 0.863 & 0.657 & 0.384 & 0.646 \\
\name{}-L & 400 & 200 & \textbf{0.919} & 0.596 & \textbf{0.975} & 0.729 & 0.566 & 0.669 \\
\name{}-H & 400 & 200 & 0.891 & \textbf{0.678} & 0.965 & \textbf{0.766} & \textbf{0.822} & \textbf{0.771} \\
\midrule
\name{}-M & 1000 & 1000 & \textbf{0.960} & 0.620 & \textbf{1.000} & \textbf{0.800} & 0.500 & 0.754 \\
\name{}-L & 1000 & 1000 & 0.940 & 0.760 & \textbf{1.000} & \textbf{0.800} & 0.875 & 0.763 \\
\name{}-H & 1000 & 1000 & \textbf{0.960} & \textbf{0.780} & \textbf{1.000} & \textbf{0.800} & \textbf{1.000} & \textbf{0.856} \\
\bottomrule
\end{tabular}

\end{table}

\begin{figure}[t]
  \centering
  \includegraphics[width=\textwidth]{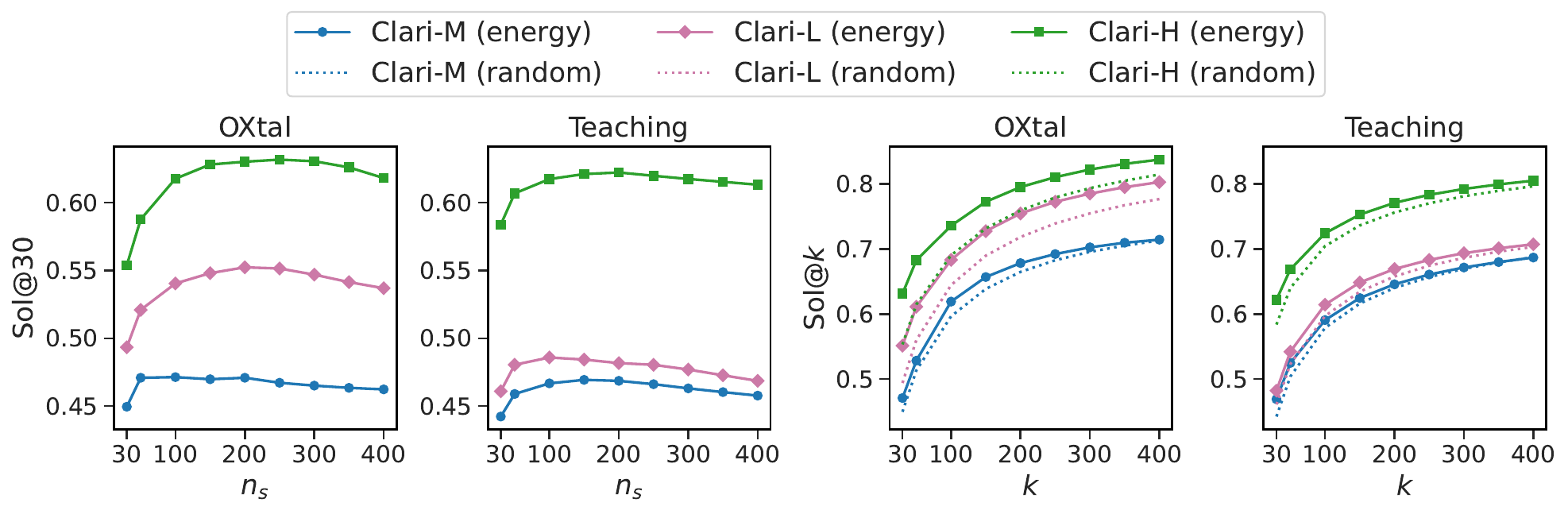}
  \caption{Inference-time scaling on OXtal's aggregated test set and the CSD Teaching Subset. \emph{Left:} Sol@30 as a function of the sampling budget $n_s$, with the top $30$ samples per target selected by UMA energy. Scaling $n_s$ does not always improve Sol@$30$, indicating ranking noise. \emph{Right:} Sol@$k$ as a function of the selection size $k$. Solid lines select top-$k$ by UMA energy with $n_s=\max(200,\,2k)$, while dotted lines draw $n_s = k$ samples. Selecting top-$k$ by UMA energy provides an advantage over random selection. Bootstrap means over 5000 resamples are plotted.}
  \label{fig:solc_vs_k}
  \vspace{-0.1in}
\end{figure}

\begin{figure}[t]
  \centering
  \includegraphics[width=0.9\textwidth]{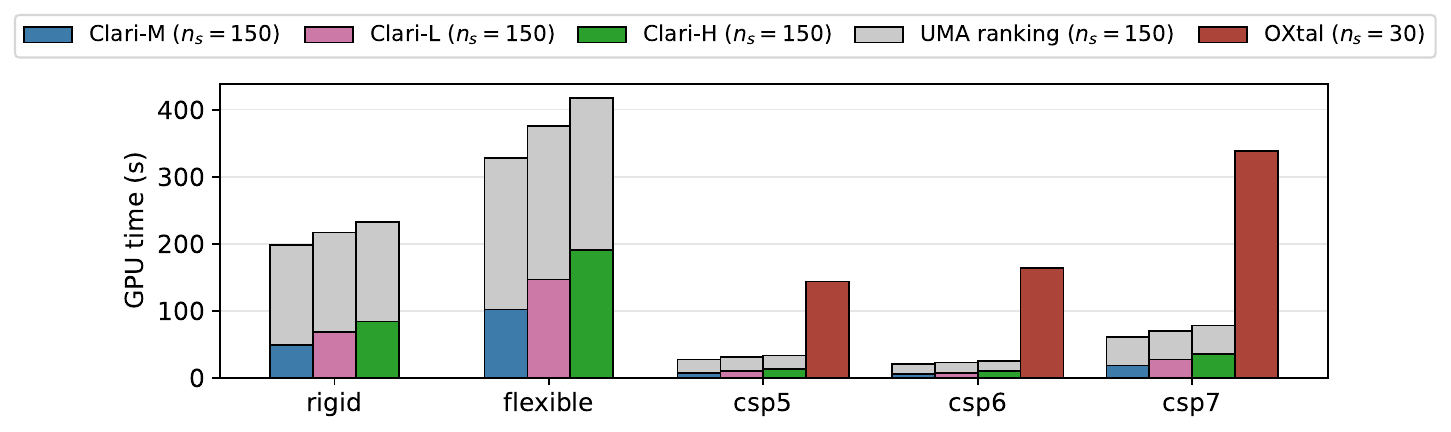}
  \caption{Per-dataset GPU wall-clock time on H100 for $n_s=150$, where we sample and rank $150$ candidates per target. Coloured bars are \name{} sampling; the grey extension is the additional cost of UMA energy ranking on the $150$ generated samples. OXtal timings (CSP5--7 only) are approximated for H100 bf16 by dividing the reported L40S timings by a factor of $2.5$.}
  \label{fig:timing}
  \vspace{-0.1in}
\end{figure}

\subsection{Comparison to baselines}\label{sec:comparison}

Leveraging the fast generation speed of \name{}, we apply inference-time scaling through best-of-$N$ sampling (Section~\ref{sec:ttscale}).
For each target crystal, we generate $n_s > k$ candidates, rank them by UMA energy without geometry optimization, and retain the $k$ lowest-energy candidates (equivalently, the top-$k$ candidates) to check Sol@$k$.
Figure~\ref{fig:solc_vs_k} shows that Sol@$k$ increases sublinearly with both the sampling budget $n_s$ and the retained set size $k$.
However, for fixed $k$, increasing $n_s$ does not always improve Sol@$k$, indicating ranking noise.
Thus, we tune $n_s$ with respect to Sol@$30$ on a held-out set of $50$ crystals drawn from the CSD Teaching Subset, minimizing leakage onto the OXtal benchmark (Figure~\ref{fig:estimate_ns}).
We find that $n_s = 150$ works best.

Table~\ref{tab:all-metrics} reports Sol across all subsets.
To estimate uncertainty, we generate a pool of $1000$ candidates per target and compute bootstrap estimates of Sol@$k$ using $5000$ resamples.
In each resample, we draw $n_s$ candidates with replacement from the $1000$-candidate pool, proceed with the above ranking procedure to compute Sol@$k$, and then average Sol@$k$ over resamples.
We report \added{bootstrap standard errors} in \Cref{tab:sol-base-all}.
The $n_s = k = 1000$ setting uses the full candidate pool and is therefore reported without uncertainty.

In the setting of $n_s=k=30$, which \added{disables UMA ranking and matches OXtal's sample count}, \name{}-M outperforms OXtal across all subsets.
\added{Increasing} the scale of \name{}-M to \name{}-L \added{to \name{}-H} provides monotonic improvements.
Inference-time scaling further improves performance, and \added{\name{}-H} with top-$30$ energy selection achieves the strongest performance among our methods on every subset\added{ except CSP5}.
When sampling $n_s = 400$ candidates and retaining the top-$200$ by UMA energy, \name{}-L\added{ and \name{}-H} surpass the average DFT participant in CSP5 (top-$464$) and CSP7 (top-$868$)\added{ despite retaining fewer candidates}.
Averaged across OXtal's test set, \added{\name{}-H takes only 2.8 seconds per molecule} to generate 150 crystal structures, or \added{6.6} seconds when downselecting to 30 with energy ranking\added{ (Figure~\ref{fig:timing})}.

\textbf{CSD Teaching Subset.} To evaluate \name{} on more realistic scenarios, we benchmark on the CSD Teaching Subset \cite{battle2010applications}, a collection of >$1000$ diverse crystals curated for educational purposes, covering a wide range of functional groups, valence-shell electron-pair repulsion (VSEPR) structure types, organometallic chemistry, flexible cycles, and stereochemistry.
We apply the same standard filters as used for training (R-factor below $10\%$, at most $512$ atoms in the unit cell), yielding $773$ crystals.
At $n_s=k=1000$, which represents how \name{} might be used in exhaustive search workflows, \name{}-L attains a Sol@$1000$ of $0.763$\added{ and \name{}-H attains $0.856$}, demonstrating \name{}'s ability to handle a wide space of chemistry.
\Cref{fig:compack} shows that \name{} can make predictions on complex molecules like metal complexes, fullerenes, and atom clusters, which are not RDKit-sanitizable and for which it would be nontrivial to produce the required input conformers for OXtal.
\added{Extended metrics are reported in Appendix~\ref{app:extended}.}

\begin{table}[t!]
\centering
\caption{Ablation of design choices on the validation set. The first row reports the quality metrics on the Cambridge Structural Database (CSD) validation set for reference. We emphasize that quality metrics are designed not to be blindly optimized but instead to provide another dimension of characterization, as even the ground truth does not obtain a perfect score.
PoseBusters is computed on the $71.3\%$ of applicable validation crystals. \name{}-M is obtained from \textbf{D} by using $\text{Beta}(1.8, 1)$ for $p(t)$ instead of the uniform distribution. Bootstrap means with standard errors are reported over 5000 resamples.}
\label{tab:ablation}
\small
\begin{tabular}{l c c c c}
\toprule
Model & Clash Rate & PoseBusters & Vol.\ Error & EMD PDD \\
\midrule
CSD (Val.) & \hphantom{0}0.80 & 92.42 & -- & -- \\
\midrule
A (Base DiT) & 25.00 $\pm$ 0.50 & 78.14 $\pm$ 0.42 & 2.07 $\pm$ 0.05 & 11.01 $\pm$ 0.05 \\
B (A + lattice tokens) & 23.25 $\pm$ 0.49 & 79.04 $\pm$ 0.41 & 2.22 $\pm$ 0.05 & 11.09 $\pm$ 0.05 \\
C (B + auxiliary losses) & 20.33 $\pm$ 0.47 & 80.11 $\pm$ 0.40 & 1.79 $\pm$ 0.04 & 10.37 $\pm$ 0.04 \\
D (C + self-cond.) & \hphantom{0}9.79 $\pm$ 0.35 & 85.23 $\pm$ 0.35 & 1.55 $\pm$ 0.04 & \hphantom{0}9.66 $\pm$ 0.03 \\
\midrule
E (D + normal $p_0$) & \hphantom{0}9.32 $\pm$ 0.33 & 84.89 $\pm$ 0.36 & 2.88 $\pm$ 0.06 & 10.71 $\pm$ 0.05 \\
\midrule
\name{}-M & \hphantom{0}9.56 $\pm$ 0.35 & \textbf{87.34 $\pm$ 0.34} & 1.59 $\pm$ 0.04 & \hphantom{0}9.56 $\pm$ 0.03 \\
\name{}-L & \textbf{\hphantom{0}7.69 $\pm$ 0.32} & 85.89 $\pm$ 0.39 & 1.50 $\pm$ 0.04 & \hphantom{0}9.28 $\pm$ 0.03 \\
\name{}-H & 10.81 $\pm$ 0.38 & 86.53 $\pm$ 0.40 & \textbf{1.39 $\pm$ 0.03} & \textbf{\hphantom{0}9.13 $\pm$ 0.03} \\
\bottomrule
\end{tabular}

\end{table}

\begin{table}[t!]
\centering
\caption{Ablation of inference discretization $(t_i)_i$ and training-time distribution $p(t)$ on the validation set. Model \textbf{D} from Table~\ref{tab:ablation} corresponds to linear discretization with uniform distribution. Bootstrap means and standard errors over 5000 resamples are reported.}
\label{tab:time}
\small
\begin{tabular}{ll c c c c}
\toprule
$(t_i)_i$ & $p(t)$ & Clash Rate & PoseBusters & Vol.\ Error & EMD PDD \\
\midrule
\multirow{4}{*}{Log} & Uniform & 11.77 $\pm$ 0.37 & 77.75 $\pm$ 0.44 & 1.55 $\pm$ 0.04 & 9.80 $\pm$ 0.03 \\
 & Logit-normal & \hphantom{0}6.84 $\pm$ 0.29 & 31.50 $\pm$ 0.34 & 1.55 $\pm$ 0.05 & 9.90 $\pm$ 0.07 \\
 & $\text{Beta}(1.8, 1)$ & 11.67 $\pm$ 0.38 & 82.46 $\pm$ 0.41 & 1.58 $\pm$ 0.04 & 9.72 $\pm$ 0.03 \\
 & Ramp & 11.26 $\pm$ 0.37 & 80.00 $\pm$ 0.42 & 1.57 $\pm$ 0.04 & 9.74 $\pm$ 0.03 \\
\midrule
\multirow{4}{*}{Linear} & Uniform & \hphantom{0}9.79 $\pm$ 0.35 & 85.23 $\pm$ 0.35 & 1.55 $\pm$ 0.04 & 9.66 $\pm$ 0.03 \\
 & Logit-normal & \oldtextbf{\hphantom{0}5.96 $\pm$ 0.27} & 75.98 $\pm$ 0.35 & \oldtextbf{1.46 $\pm$ 0.04} & 9.56 $\pm$ 0.05 \\
 & $\text{Beta}(1.8, 1)$ & \hphantom{0}9.56 $\pm$ 0.35 & \oldtextbf{87.34 $\pm$ 0.34} & 1.59 $\pm$ 0.04 & 9.56 $\pm$ 0.03 \\
 & Ramp & \hphantom{0}8.91 $\pm$ 0.33 & 86.08 $\pm$ 0.35 & 1.53 $\pm$ 0.04 & \oldtextbf{9.55 $\pm$ 0.03} \\
\bottomrule
\end{tabular}

\end{table}

\subsection{Ablations}
\label{sec:ablations}

\added{To better understand the performance of \name{}, we ablate key design choices on the validation set.} We generate $20$ samples per crystal and report bootstrap means over 5000 resamples. Each resample draws $5$ values per crystal with replacement, aggregates them into a per-crystal score (using a mean for quality metrics and a minimum for reconstruction), and then averages across crystals.
Table~\ref{tab:ablation} provides our main results, whereas Table~\ref{tab:time} provides ablations across time distribution and discretization. \added{Apart from the scaled-up \name{}-L and \name{}-H, all models in this section use the hyperparameters of \name{}-M, which are listed in Appendix~\ref{app:model:hparams}.}

\textbf{Architecture and loss.} In \textbf{A}, we start with a standard flow matching model with a DiT architecture that predicts lattice vectors from mean-pooled atom features. Then, in \textbf{B}, we \added{instead treat} the lattice as three atom-level tokens and find that it is roughly quality-neutral, but simplifies the architecture to a single token stream and qualitatively stabilizes optimization (Figure~\ref{fig:AB-loss-curve}).
Adding the auxiliary volume and pairwise distance losses (\textbf{C}) and self-conditioning (\textbf{D}) results in notable improvements across all metrics.

\textbf{Source distribution.} Models \textbf{A}\added{--}\textbf{D} use a fitted lattice prior that closely matches the training distribution (Figure~\ref{fig:pt_source}), as discussed in Section~\ref{sec:flow}. In \textbf{E}, we test the standard normal prior and find it to be detrimental, \added{degrading reconstruction metrics to around the level of \textbf{B}, with an even larger volume error}. The clash rate is the only metric that improves (marginally), though this may be due to the model generating unrealistically large and sparse unit cells, as evidenced by the degraded EMD PDD.  

\textbf{Timestep settings.}
Flow matching is sensitive to both the training-time distribution $p(t)$ and inference discretization $(t_i)_i$.
Models for protein and small-molecule generation tend to skew both toward the late-time regime, since it governs local detail \cite{geffner2025proteina,vonessen2025tabasco,irwin2025semlaflow}. 
In Table~\ref{tab:time}, we explore such settings for our crystal domain. \added{Contrary to prior work, we find log discretization consistently harmful, most notably degrading} the clash rate and EMD PDD. This supports our hypothesis that crystals are not purely local: global packing geometry is established at intermediate times, so over-emphasizing $t \to 1$ risks starving these stages. Indeed, a logit-normal distribution that focuses mass around the mid-range improves packing metrics over the uniform distribution. However, it also degrades the PoseBusters quality (in fact, collapsing when combined with log discretization), suggesting that \added{underweighting late times} can also be problematic. Conversely, a $\text{Beta}(1.8, 1)$ \cite{vonessen2025tabasco} distribution shows better PoseBusters validity but performs worse otherwise. Finally, we demonstrate that we can successfully trade off between these endpoints through a \textit{ramp} distribution that places the majority of its mass uniformly on $[0.5, 1]$ (Figure~\ref{fig:pt_source}\added{a}). However, its performance is relatively similar to the beta distribution, so our final models use $\text{Beta}(1.8, 1)$ and we leave further exploration to future work.   

\textbf{Model scale.} Scaling \name{}-M from $88$M to \name{}-L at $173$M parameters results in a general improvement in reconstruction, which \added{becomes more apparent on} the test set. Surprisingly, \name{}-L trains only $\sim$20\% slower than \name{}-M, despite having nearly $2\times$ as many parameters. \added{Scaling further to \name{}-H at $350$M parameters continues to improve reconstruction metrics, although the clash rate increases.}

\section{Conclusion}
\label{sec:conclusion}

We present \name{}, a flow matching model for organic crystal structure prediction that operates directly on a single unit cell using a pair-bias DiT, avoiding both bulk expansion and triangle-update layers.
These architectural refinements improve prediction quality while making sampling roughly $15$--$30\times$ faster than OXtal on the CSP Blind Tests ($5$--$8\times$ end-to-end when UMA energy ranking is included).
By jointly modelling heavy atoms and hydrogens, \name{} produces structures amenable to direct energy-based ranking, enabling inference-time scaling through best-of-$N$ sampling without decoration or relaxation.
We further analyze the sources of these gains through ablations and introduce the CSD Teaching Subset, a benchmark spanning chemically complex systems that are excluded from or underrepresented in prior evaluations.

Our results show that explicit lattice modelling provides an efficient route to generative CSP.
Compared with bulk representations, modelling a single unit cell avoids redundancy and yields substantial computational savings, reducing crystal generation from minutes to seconds.
This speed unlocks a different use case for CSP: in the case of screening large virtual libraries of millions of candidate molecules for solid-state properties \cite{ishii2020charge}, it may be desirable to simply obtain a \emph{decent} crystal structure within a few-second turnaround time.
While prior methods can be accelerated with few-step generation methods \cite{consistency, boffi2026how, geng2026mean, sabour2026align, deng2026generative}, these techniques are orthogonal to and fully compatible with \name{}'s architectural speedups.

At the same time, we view explicit lattice modelling as complementary to bulk approaches such as OXtal.
While the unit cell representation is almost always more compact, lattice-free modelling may better accommodate amorphous systems \cite{cordova2021structure}, where the unit cell becomes extremely large or ill-defined.

\textbf{Limitations and future work.}  \added{Search-based CSP offers high accuracy and recall at slow speed, whereas generative CSP offers decent accuracy and recall at fast speed. \name{}'s capabilities map onto virtual screening, where high throughput is necessary but accuracy and recall do not need to be watertight, rather than pharmaceutical polymorph screening, where both are necessary and speed is secondary. We do not advocate that \name{} should replace traditional CSP pipelines, but it may supplement them with additional crystal candidates. In preliminary polymorph coverage experiments (Appendix~\ref{app:polymorphs}), \name{}-H recovers $61$--$75\%$ of the observed polymorphs in OXtal test families using only $150$ samples per family, though coverage is weaker on the flexible and blind-test families.}
On the modelling front, \name{} conditions on a 2D molecular graph that is agnostic to stereochemistry, \added{a key limitation for pharmaceutical applications}.
Conditioning on an explicit 3D conformer or augmenting the graph with chiral tags would allow for more controllable generation.
In addition, \name{} requires as input the number of copies of each molecule in the unit cell, which may not always be known at \added{inference time}, although it is straightforward to sweep over common values of $Z=4,2,1$.
\added{On the data side, we contribute a processed CSD dataset with held-out splits and a new CSD Teaching Subset benchmark, but both leave room for improvement.}
Constructing test splits that cleanly probe generalization is harder for crystals than for molecules, since graph-based similarity measures are less established for multi-component crystals that RDKit cannot sanitize.
Our splits combine refcode grouping with heuristic component-level filtering as a step towards this, but more principled splitting protocols would be valuable for future work.
In addition, evaluating generative CSP at scale is hindered by the lack of fast and robust crystal-similarity metrics.
We note that Sol is an approximate metric \added{requiring only 8 of 15 molecules to match}, which limits the scope of our results.
We also observe that \name{} occasionally generates crystals with steric clashes or unphysical voids.
Reward alignment or inference-time steering \cite{singhal2025a,skreta2025feynmankac,potaptchik2026meta} can help reduce these errors.
Another promising direction is to apply inference-time steering towards a given powder diffraction pattern \cite{li2025powder}, space group \cite{watson2023novo}, energy, or density.
We hope \name{} encourages a reassessment of architectural choices inherited from biomolecular generative models, and enables virtual screening of solid-state properties for pharmaceuticals, agrochemicals, and organic semiconductors.

\textbf{Broader Impacts.}
Our work accelerates crystal structure prediction for applications in pharmaceuticals and organic materials. As a dual-use consideration, improved prediction may enable more efficient screening of energetic materials, including the identification of more stable or higher-density explosives. 
Improved structural understanding may also inform safer formulation and handling practices in industrial uses of energetic materials (e.g., mining and construction). While practical deployment of new energetic materials remains constrained by challenges in synthesis and experimental validation, methods such as \name{} may lower computational barriers to discovery. We therefore emphasize responsible use within established safety and regulatory frameworks.

\section*{Acknowledgments}
The authors thank Matthew Spellings, Michael Kilgour, and Olivier Trottier for helpful discussions.
A.H.C. acknowledges the generous support of the Canada 150 Research Chairs program through A.A.-G.
L.M. acknowledges support from the Prof. Dr. Sc. Jasna \v{S}imuni\'{c}-Hrvoi\'{c} Foundation Fellowship.
A.A.-G. thanks Anders~G.~Fr{\o}seth for his generous support. A.A.-G. also acknowledges the generous support of Natural Resources Canada and the Canada 150 Research Chairs program.
\acknowAC
\acknowGEN{SciNet HPC Consortium (\url{https://scinethpc.ca/})}
\acknowSciNet{Trillium}
\acknowDARPA

\bibliographystyle{unsrtnat}
\bibliography{refs}

\newpage

\appendix

\section{Dataset}
\label{app:dataset}

\subsection{Crystal processing pipeline}

The Cambridge Structural Database (CSD) requires a license. We obtained an academic license via the University of Toronto.

We download CSD entry metadata using the CSD Python API and the raw CIF and MOL2 files using ConQuest. Given a CSD entry, the following workflow is performed:
\begin{enumerate}
    \item Metadata filtering according to the criteria discussed in Section \ref{sec:experimental-setup}.
    \item Parsing the crystal's asymmetric unit from its CIF and MOL2 file.
    \item Caching component isomorphisms within the asymmetric unit.
    \item Expanding the asymmetric unit to the full unit cell by applying space group operations.
    \item Distance-based filtering to remove polymers and steric clashes.
    \item Unit cell featurization (e.g., SMILES for splitting on test components).
\end{enumerate}
Crystals that are successfully processed without error are aggregated and split into training, validation, and test datasets. We elaborate on the critical steps \added{3} and 5.

\textbf{Isomorphism caching.} For crystal alignment (Section~\added{\ref{sec:augment}}), it is useful to know which components in the unit cell are isomorphic, and if so, what those isomorphisms are. We cache these during our data processing pipeline to avoid expensive isomorphism checks during training. To further improve efficiency, we compute these on the asymmetric unit, which generally has many fewer components than the full unit cell. When the asymmetric unit is replicated in step (4), these isomorphisms can then be easily extended. Note that we do not attempt to intractably enumerate all isomorphisms between components but rather find a single one, if it exists.

\begin{figure}[th]
\centering
\includegraphics[width=\textwidth]{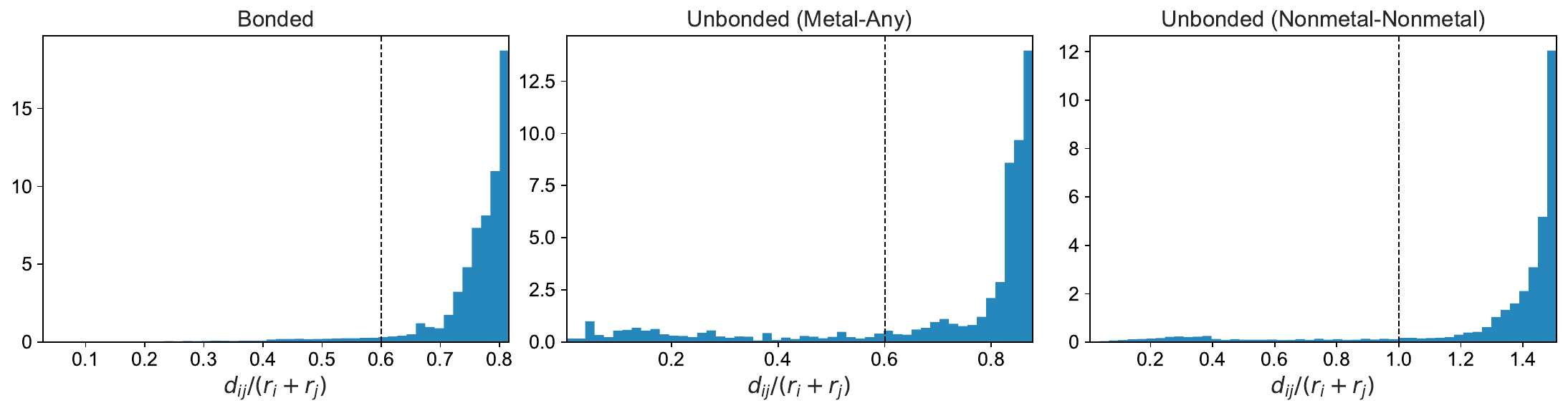}
\caption{Distribution of normalized periodic interatomic distances $d_\mL(\vp_1, \vp_2) / (r(a_1) + r(a_2))$ across the CSD, with the threshold $\alpha$ (Equation \ref{eq:dist-lbound}) shown in a dashed line. For each atom in the CSD, we gather distances of adjacent bonded atoms (left) and the nearest unbonded atoms. The latter case is further broken down based on whether one of the atoms is a metal (middle) or both are nonmetals (right). Histograms are each truncated at their lower $0.5\%$ tail.}
\label{fig:csd_distances}
\end{figure}

\textbf{Distance-based filtering.} We more strictly filter out polymeric or noisy crystals (e.g., due to improper disorder specification) by checking that all atoms are sufficiently distanced. Let $M_1$ and $M_2$ be two (not necessarily distinct) components obtained after step (4). Let $\vp_1$ be the position of an atom $a_1$ from $M_1$, and similarly for $\vp_2$ and $a_2$ from $M_2$. We check that
\begin{equation}\label{eq:dist-lbound}
d_\mL(\vp_1, \vp_2) \geq \alpha\left(r(a_1) + r(a_2)\right), \quad
\text{where }
\alpha = \begin{cases}
0.6, & \text{if $a_1$ or $a_2$ is a metal},\\
0.6, & \text{if $a_1$ and $a_2$ are bonded}, \\
1.0, & \text{otherwise},
\end{cases}
\end{equation}
where $\mL$ is the cell lattice and $r(\cdot)$ is the CSD covalent radius (\si{\angstrom}) \cite{alvarez2008covalent, ccdcradii}. Threshold distances $\alpha$ were chosen based on the periodic distance distribution of CSD (Figure \ref{fig:csd_distances}). A minor technicality is that two distinct but exactly superimposable components can arise, for example, when a component in the asymmetric unit is invariant under the space group. In this case, we should discard one component rather than rejecting the entire CSD entry. If $M_1 \neq M_2$ are isomorphic but clashing, we can check for this case by performing an optimal atomic assignment \cite{kuhn1955hungarian} and checking that all atoms are within \SI{0.01}{\angstrom} deviation.

\subsection{Dataset splitting}

The OXtal test crystal families and the CSD Teaching Subset are combined to create our held-out test pool. To mitigate leakage, we exclude entire 6-letter refcode families of test entries and additionally exclude crystals sharing an RDKit-sanitizable component with any test structure. The component must be RDKit sanitizable so we can perform the equality check using its canonical SMILES rather than expensive graph isomorphism checks. Moreover, the component must also have over $7$ heavy atoms, so that ubiquitous but small molecules such as water or hexafluorophosphate are not considered.
The validation set is created by holding out 1000 refcode families from the remainder.
The final split counts, before benchmark-specific evaluation filters, contain 917,014 training, 1048 validation, and 2996 test examples across 859,866, 1000, and $919$ families, respectively; the test count includes all held-out crystals, including those with more than $512$ atoms in the unit cell.

\section{Modelling}
\label{app:model}

\subsection{Source distribution}
\label{app:model:source}

We decompose lattice matrices $\mL = (\vl_1, \vl_2, \vl_3)^\top \in \R^{3 \times 3}$ of an $N$-atom unit cell into three components and model each as independent Gaussians fitted to the CSD training set:
\begin{itemize}
    \item The atom density $\rho  \sim \mathcal{N}(\mu_\rho, \sigma_\rho^2)$, where $\rho = N/V$ for $V = |\det \mL|$.
    \item The cell angles $\alpha, \beta, \gamma \sim \mathcal{N}(\mu_{\circ}, \sigma_{\circ}^2)$ i.i.d., where $\alpha = \mathrm{angle}(\vl_1, \vl_2)$ and similarly for $\beta, \gamma$.
    \item The normalized cell lengths $(a, b, c) \sim \mathcal{N}(\bm{\mu}_\ell, \bm{\Sigma}_\ell)$, where $a = ||\vl_1||_2/V^{1/3}$ and $b, c$ are defined similarly, sorted such that $a \leq b \leq c$. Note that $\bm{\Sigma}_\ell$ is not diagonal.
\end{itemize}
Given a sample $(\rho, \alpha, \beta, \gamma, a, b, c)$, we can reconstruct a matrix $\mL$ that adheres to the given parameters. To randomize the pose of the matrix, we finally apply a random signed permutation and rotation.

\subsection{Architecture}
\label{app:model:architecture}

Let $\mathrm{Linear}(\cdot)$ and $\mathrm{LN}(\cdot)$ denote a linear layer and LayerNorm. Let
\begin{align}
    \mathrm{Mod}(\vx, \vc) &= \mathrm{Linear}(\vc) \odot \vx + \mathrm{Linear}(\vc),\\
    \mathrm{Gate}(\vx, \vc) &= \mathrm{Linear}(\vc) \odot \vx, \\
    \mathrm{AdaLN}(\vx, \vc) &= \mathrm{Mod}(\mathrm{LN}(\vx), \vc), \\
    \mathrm{Transition}(\vx) &= \mathrm{Linear}(\mathrm{SwiGLU}(\mathrm{Linear}(\vx))),
\end{align}
and let $\mathrm{PairAttention}(\vh, \vz)$ denote attention with pairwise bias via $\mathrm{Linear}(\vz)$. The architecture of \name{} is summarized in Algorithm~\ref{alg:dit}. The input features to \name{} are listed in Table~\ref{tab:features}.

\begin{algorithm}[h]
\caption{\name{} architecture.}
\label{alg:dit}
\begin{algorithmic}[1]
\Require Sequence dimension $d$, pair dimension $d_z$, conditioning dimension $d_c$, trunk depth $D$, number of attention heads $H$ (Table~\ref{tab:hparams}).
\Statex {\small \textcolor{algcolor}{Create conditioning features}}
\State $\vc \gets$ embed conditioning features (Table~\ref{tab:features}) \Comment{$\vc \in \sR^{d_c}$}
\For{ $i = 1, 2$}
\State $\vc \gets \vc + \mathrm{Transition}(\mathrm{LN}(\vc))$
\EndFor
\State $\vc \gets \mathrm{LN}(\vc)$
\Statex {\small \textcolor{algcolor}{Create sequence features}}
\State $\vh \gets$ embed sequence features (Table~\ref{tab:features}) \Comment{$\vh \in \sR^{n \times d}$}
\State $\vh \gets \mathrm{Mod}(\mathrm{tanh}(\vh), \vc)$
\State $(\vh, \vu, \vv) \gets \mathrm{Transition}(\vh)$ \Comment{split as $d \oplus d_z \oplus d_z$}
\Statex  {\small \textcolor{algcolor}{Create pair features}}
\State $\vz \gets$ embed pair features (Table~\ref{tab:features}) \Comment{$\vz \in \sR^{n \times n \times d_z}$}
\State $\vz \gets \vz + (\vu_i + \vv_j)_{ij}$
\State $\vz \gets \mathrm{Mod}(\mathrm{tanh}(\vz), \vc)$
\Statex {\small \textcolor{algcolor}{Main trunk}}
\For{$\ell = 1, \dots, D$}
    \State $\vh \gets \vh + \mathrm{Gate}\!\left(\mathrm{PairAttention}\!\left(\mathrm{AdaLN}(\vh, \vc), \vz\right), \vc\right)$
    \State $\vh \gets \vh + \mathrm{Gate}\!\left(\mathrm{Transition}\!\left(\mathrm{AdaLN}(\vh, \vc)\right), \vc\right)$
\EndFor
\State $\vv \gets \mathrm{Transition}\!\left(\mathrm{AdaLN}(\vh, \vc)\right)$ \Comment{ $\vv \in \sR^{n \times 3}$}
\State \Return $\vv$
\end{algorithmic}
\end{algorithm}

\begin{table}[h]
\centering
\caption{Input features consumed by \name{}. Within each stream, we take the sum of the individual embeddings as the final sequence, pair, or conditioning features.}
\label{tab:features}
\renewcommand{\arraystretch}{1.15}
\begin{tabularx}{\textwidth}{llX}
\toprule
 & Input & Featurization \\
\midrule
Seq.
 & Cartesian $\vx_t$  & Linear and sinusoidal.  \\
 & Fractional $\vx_t$        & Linear and sinusoidal (periodic in $1$). \\
 & Self-cond. $\hat{\vx}_1$ &  Linear and sinusoidal. \\
 & Element          &  Embedding. Uncommon elements map to a $*$ type. Lattice nodes are treated as 3 new elements. \\
 & Atomic charges & Linear  and embedding with bins $\{\shortminus 2\text{-}, \shortminus 1, 0, 1, 2\text{+}\}$. \\
 & Atomic degrees & Linear and embedding with bins $\{0, \ldots, 9\text{+}\}$. \\
 & Atomic radii           & Linear  of covalent and VDW radii. \\
 & Adjacent bonds & Linear projection of binary indicator vector.  \\
\midrule
Pair
 & Bonds & Embedding. \\
 & Topological dist.\  & Embedding with bins $\{0, \ldots, 15, 16\text{+}, \infty\}$. \\
 & Cartesian dist.\    & Embedding with 128 bins over $[0, 32]$\,\AA{}. \\
 & Periodic dist.\    & Embedding with 128 bins over $[0, 32]$\,\AA{}. \\
\midrule
Cond.
& Timestep $t$               & Sinusoidal. \\
& Lattice  $\mL$           & Linear projection of $\mL$, $\mL^{-1}$, $\mL^\top\mL$, and  $\det\mL$. \\
& Formula            & Linear projection of count vector. \\
& Self-cond.\  & Embedding (binary) for whether $\hat{\vx}_1 = \varnothing$. \\
\bottomrule
\end{tabularx}
\end{table}

\subsection{Inference batching}
The dataloader yields one crystal per batch; we then replicate it $B(n)$ times in the batch dimension for parallel sampling, where $B(n) = 1000, 500, 200, 25, 1$ for $n < 200, < 300, < 500, \leq 1000, > 1000$ atoms. On an out-of-memory error, $B$ is halved and the chunk is retried. Before beginning timing, we warm up compilation with dummy runs on the first $5$ batches.

\subsection{Auxiliary losses}
\label{app:model:losses}

The volume and pair losses are applied between the one-step estimate $\hat{\vx}_1 = \vx_t + (1-t)\, v_\theta(\vx_t, t)$ and ground truth $\vx_1$. The auxiliary volume loss is the relative error of the predicted volume:
\begin{equation}\label{eq:loss-vol}
    \Ls_{\mathrm{vol}} = \left| \frac{|\det \hat{\mL}_1|}{|\det \mL_1|} - 1 \right|,
\end{equation}
where $\hat{\mL}_1$ and $\mL_1$ are the lattice matrices obtained from $\hat{\vx}_1$ and $\vx_1$, respectively. Now, let $\hat{d}_{ij}$ and $d_{ij}$ be the periodic distance (\si{\angstrom}) between the $i$-th and $j$-th atoms in $\hat{\vx}_1$ and $\vx_1$, respectively, and $\alpha_{ij}$ be the distance threshold from Equation~\ref{eq:dist-lbound}.
The auxiliary pair loss is
\begin{equation}
    \Ls_{\mathrm{pair}} = \sum_{(i, j) \in \Lambda}\!\left[ \bigl| \hat d_{ij} - d_{ij} \bigr| + 5 \cdot \max\left(0, \alpha_{ij} - \hat{d}_{ij}\right)  \right],
\end{equation}
where $\Lambda = \{i \neq j \mid d_{ij} < 15\,\text{\AA} \text{ or } \hat{d}_{ij} < \alpha_{ij} \}$. The pair loss combines a distance matching term with a clash term. To compute the periodic distances tractably and stably, we use the \added{estimate}:
\[
d_{\mL}(\vp_i, \vp_j) \leq \min_{\vz \in \{-1, 0, 1\}^3}\left\lVert\vp_i - \vp_j - \mL^\top\vz\right\rVert_2,
\]
which tends to be exact for most crystals.

\subsection{Optimal-transport coupling}
\label{app:model:ot}

We approximately align $\vx_1$ to $\vx_0$ before constructing the interpolant $\vx_t$ by composing three tractable substeps. First, we align the pair purely on their lattices to obtain a good initial pose. Then, we iterate between permutation and pose alignment twice. Let $(\mL_1, \mC_1)$ and $(\mL_0, \mC_0)$ be the lattice and atomic rows of $\vx_1$ and $\vx_0$, respectively. The steps are now described:

\textbf{Exact lattice alignment.} The goal is to find a signed permutation $\mPi$ and rotation $\mR$ that minimize $||\mPi\mL_1\mR^\top - \mL_0||_F$. We can exactly solve this joint problem by enumerating over all such $\mPi$, and for each, finding the optimal $\mR$ using the Kabsch algorithm \cite{kabsch1976solution}.

\textbf{Permutation alignment.}
With pose fixed, the goal is to permute the rows of $\vx_1$ to minimize its distance to $\vx_0$. The lattice and atomic rows can be handled independently. For the former, we can again brute-force search across all signed permutations. For the latter, we need to find a permutation $\mPi$ that minimizes $||\mPi\mC_1 - \mC_0||_F$ such that $\mPi$ is also an automorphism of the crystal molecular graph. Since enumerating all such automorphisms is intractable, we use the isomorphisms cached from our dataset processing (Appendix \ref{app:dataset}) to get an approximate answer. Specifically, for all isomorphic components between $\mC_1$ and $\mC_0$, we compute their RMSD using the atom assignment given by the cached isomorphism. Then, we resolve the assignment of bodies in $\vx_1$ to those in $\vx_0$ using the Hungarian algorithm \cite{kuhn1955hungarian}. Note that this does not search the full space because we only consider one of many possible isomorphisms between components.

\textbf{Pose alignment.}
With order fixed, we can $\mathrm{SO}(3)$-align $\vx_1$ to $\vx_0$ using a weighted Kabsch algorithm in which the three lattice rows carry equal weight to the $N$ atom rows.

\subsection{Hyperparameters}
\label{app:model:hparams}

Table~\ref{tab:hparams} lists the architectural and training hyperparameters for the medium\added{, large, and huge} \name{} models. We use a feedforward expansion of $2.66\times$ to match the parameters of a standard 4$\times$ non-GLU feedforward\added{, except for \name{}-H, which uses $4\times$}. We train the Transformer weight matrices using Muon \cite{jordan2024muon, liu2025muon} and all other parameters using Adam \cite{kingma2015adam}.
We also maintain an exponential moving average (EMA) of all parameters for sampling and evaluation.

\begin{table}[h]
\centering
\caption{Training and architectural hyperparameters for \name{}.}
\label{tab:hparams}
\begin{tabular}{lccc}
\toprule
 & \name{}-M  & \name{}-L & \added{\name{}-H} \\
 Parameters & 88M & 173M & \added{350M} \\
\midrule
Sequence dim $d$            & $512$           & $768$ & \added{$1024$} \\
Pair dim $d_z$              & $64$            & $64$ & \added{$64$} \\
Conditioning dim $d_c$      & $512$           & $512$ & \added{$512$} \\
Depth $D$                   & $16$            & $16$ & \added{$16$} \\
Heads $H$                   & $8$             & $12$ & \added{$16$} \\
Feedforward expansion       & $2.66\times$       & $2.66\times$ & \added{$4\times$} \\
\midrule
Training steps                 & 150,000     & 150,000 & \added{150,000} \\
Effective batch size        & $128$           & $256$ & \added{$512$} \\
Optimizer & Muon, Adam            & Muon, Adam & \added{Muon, Adam} \\
Learning rate               & $0.0005$ & $0.0005$ & \added{$0.0005$} \\
Warmup steps                & 5000       & 5000 & \added{5000} \\
Weight decay & 0 & 0 & \added{0} \\
EMA decay                   & $0.999$         & $0.999$ & \added{$0.999$} \\
\midrule
GPUs                        & $4\times$ H100  & $8\times$ H100 & \added{$16\times$ H100} \\
Training time                  & 14 h   & 17 h & \added{21 h} \\
\bottomrule
\end{tabular}
\end{table}

\section{Metrics}
\label{app:metrics}

We consider \emph{quality} metrics, assessing the inherent plausibility of a single sample, and \emph{reconstruction} metrics, measuring agreement with a reference. Clash rate and PoseBusters are quality metrics while the others are reconstruction metrics.

\textbf{Clash rate.} We consider two \textit{distinct} components within a crystal clashing if an atom from each has periodic distance less than their sum of covalent radii (i.e., Equation \ref{eq:dist-lbound} with $\alpha = 1$). We report the fraction of crystals with at least one pair of clashing components.

\textbf{PoseBusters.} We report the average PoseBusters \cite{buttenschoen2024posebusters} validity of all components in the unit cell, following the settings from \citet{vonessen2025tabasco}. Since PoseBusters relies on RDKit, we process the predicted crystal by disconnecting organometallic bonds and then excluding any components that (i) have no bonds, (ii) contain an element outside of H, C, N, O, F, P, S, Cl, Br, I, (iii) are not RDKit sanitizable, (iv) contain added hydrogens after sanitization. That is, PoseBusters is run only on a subset of applicable components. This metric is averaged over all crystals that have at least one applicable component.

\textbf{Relative volume error.} This is $\Ls_{\mathrm{vol}}$ in Equation~\ref{eq:loss-vol}.

\textbf{EMD PDD.}
The pointwise distance distribution (PDD) \cite{widdowson2022resolving} is an isometry invariant of periodic point sets, computed from sorted pairwise interatomic distances of the nearest $k = 100$ neighbours to each point.
We report the ($L_1$) earth mover's distance between the PDDs of the predicted and reference crystals.

\textbf{COMPACK \text{RMSD}$_{15}$.}
COMPACK \cite{chisholm2005compack} matches a 15-molecule cluster between the predicted and reference packings, ignoring hydrogens and all bond information so that the comparison is sensitive only to heavy-atom packing.
For fair comparison we inherit the OXtal configuration verbatim: $50\%$ distance tolerance and $75^{\circ}$ angle tolerance.
These tolerances are loose, but we keep them unchanged so that all reported numbers are directly comparable to OXtal. We also only consider $\text{RMSD}_{15} < 2.0$\,\AA{}.
Samples clash if any inter-body atom pair sits at a distance less than the sum of their van der Waals radii minus $0.7$\,\AA{}.
A target is counted as solved (Sol) if at least one of the $k$ generated candidates obtains a COMPACK match rate of at least 8 out of 15 molecules to the experimental structure with $\text{RMSD}_{15} < 2.0$\,\AA\ and no detected collisions.
OXtal additionally defines \added{collision, packing-similarity, and conformer-recovery metrics. We focus on Sol since it is the critical metric for downstream applications, and report the extended metrics in Table~\ref{tab:oxtal-extended}.}
COMPACK is computationally expensive and is therefore reserved for the test set.

\textbf{OXtal comparison protocol.}
For direct comparison we adopt the OXtal test set of $119$ crystals: $50$ rigid, $50$ flexible, and the CSP blind-test entries ($6$ from CSP5, $5$ from CSP6, and $8$ from CSP7).
Model inputs are taken from the CSD entry with the lowest R-factor\added{, i.e., molecular graphs and multiplicities}. When a crystal is associated with multiple CSD refcodes of the same family, we evaluate against the three entries with the lowest R-factors and report the best COMPACK match. For the Teaching Subset, we only match against the directly corresponding refcode.
Generated structures are evaluated as produced by the model, with no relaxation, energy minimization, or DFT polish. The only post-processing is the UMA-based ranking described in Section~\ref{sec:ttscale}.

\section{Additional results}

\added{\subsection{Extended baseline comparisons}
\label{app:extended}}

\added{Tables~\ref{tab:sol-strict-all} and \ref{tab:sol-base-all} report the Sol@$k$ results of Table~\ref{tab:all-metrics} with standard errors under the strict $15/15$ and $8/15$ molecule match requirements, respectively. Table~\ref{tab:oxtal-extended} reports the collision (Col$_S$), packing similarity (Pac$_S$, Pac$_C$), and conformer recovery (Rec$_S$, Rec$_C$) metrics of \citet{jin2025oxtal}, where $S$ and $C$ denote sample-level and crystal-level aggregation. \name{} shows no collisions and outperforms OXtal on conformer recovery. \name{} appears to underperform on packing similarity, which we believe is due to differences between bulk and unit cell methods: a unit cell repetitively tiles space and so offers fewer packing environments to match against, whereas a bulk crop gives OXtal samples more chances to find matching clusters, even if they ultimately do not pass the RMSD threshold needed for Sol.}

\begin{table}[!htbp]
\centering
\small
\caption{The Sol@$k$ metrics from Table \ref{tab:all-metrics}, except with a \added{stricter} $15$-molecule match requirement. Bootstrap
means and standard errors over 5000 resamples are reported, where appropriate.}
\label{tab:sol-strict-all}
\begin{tabular}{lrr c c c}
\toprule
 & & & Rigid & Flexible & Teach. \\
Method & $n_s$ & $k$ & (50) & (50) & (773) \\
\midrule
\name{}-M & 30 & 30 & 0.288 $\pm$ 0.040 & 0.037 $\pm$ 0.024 & 0.140 $\pm$ 0.008 \\
\name{}-M & 150 & 30 & 0.344 $\pm$ 0.043 & 0.045 $\pm$ 0.025 & 0.168 $\pm$ 0.008 \\
\name{}-L & 30 & 30 & 0.373 $\pm$ 0.043 & 0.071 $\pm$ 0.031 & 0.188 $\pm$ 0.009 \\
\name{}-L & 150 & 30 & 0.435 $\pm$ 0.045 & 0.131 $\pm$ 0.036 & 0.223 $\pm$ 0.009 \\
\name{}-H & 30 & 30 & 0.461 $\pm$ 0.044 & 0.131 $\pm$ 0.037 & 0.292 $\pm$ 0.009 \\
\name{}-H & 150 & 30 & \textbf{0.539 $\pm$ 0.039} & \textbf{0.208 $\pm$ 0.039} & \textbf{0.345 $\pm$ 0.009} \\
\midrule
\name{}-M & 400 & 200 & 0.581 $\pm$ 0.040 & 0.189 $\pm$ 0.037 & 0.316 $\pm$ 0.008 \\
\name{}-L & 400 & 200 & 0.684 $\pm$ 0.038 & 0.294 $\pm$ 0.039 & 0.389 $\pm$ 0.008 \\
\name{}-H & 400 & 200 & \textbf{0.724 $\pm$ 0.028} & \textbf{0.396 $\pm$ 0.036} & \textbf{0.518 $\pm$ 0.009} \\
\midrule
\name{}-M & 1000 & 1000 & 0.720 & 0.380 & 0.446 \\
\name{}-L & 1000 & 1000 & \textbf{0.840} & 0.460 & 0.524 \\
\name{}-H & 1000 & 1000 & 0.800 & \textbf{0.520} & \textbf{0.655} \\
\bottomrule
\end{tabular}

\vspace{\baselineskip}

\begin{tabular}{lrr c c c}
\toprule
 & & & CSP5 & CSP6 & CSP7 \\
Method & $n_s$ & $k$ & (6) & (5) & (8) \\
\midrule
\name{}-M & 30 & 30 & 0.055 $\pm$ 0.090 & 0.007 $\pm$ 0.036 & 0.101 $\pm$ 0.060 \\
\name{}-M & 150 & 30 & 0.065 $\pm$ 0.092 & 0.027 $\pm$ 0.069 & 0.106 $\pm$ 0.045 \\
\name{}-L & 30 & 30 & 0.187 $\pm$ 0.139 & 0.070 $\pm$ 0.109 & 0.121 $\pm$ 0.021 \\
\name{}-L & 150 & 30 & 0.305 $\pm$ 0.159 & \textbf{0.155 $\pm$ 0.136} & 0.122 $\pm$ 0.020 \\
\name{}-H & 30 & 30 & 0.333 $\pm$ 0.141 & 0.075 $\pm$ 0.110 & 0.132 $\pm$ 0.032 \\
\name{}-H & 150 & 30 & \textbf{0.480 $\pm$ 0.141} & 0.141 $\pm$ 0.132 & \textbf{0.143 $\pm$ 0.044} \\
\midrule
\name{}-M & 400 & 200 & 0.331 $\pm$ 0.129 & 0.066 $\pm$ 0.094 & 0.145 $\pm$ 0.046 \\
\name{}-L & 400 & 200 & 0.602 $\pm$ 0.154 & \textbf{0.399 $\pm$ 0.139} & 0.125 $\pm$ 0.000 \\
\name{}-H & 400 & 200 & \textbf{0.748 $\pm$ 0.093} & 0.300 $\pm$ 0.108 & \textbf{0.205 $\pm$ 0.080} \\
\midrule
\name{}-M & 1000 & 1000 & 0.500 & 0.200 & 0.250 \\
\name{}-L & 1000 & 1000 & \textbf{0.833} & \textbf{0.600} & 0.125 \\
\name{}-H & 1000 & 1000 & \textbf{0.833} & \textbf{0.600} & \textbf{0.375} \\
\bottomrule
\end{tabular}

\end{table}

\begin{table}[!htbp]
\centering
\small
\caption{The Sol@$k$ metrics from Table \ref{tab:all-metrics}. Bootstrap
means and standard errors over 5000 resamples are reported, where appropriate.}
\label{tab:sol-base-all}
\begin{tabular}{lrr c c c}
\toprule
 & & & Rigid & Flexible & Teach. \\
Method & $n_s$ & $k$ & (50) & (50) & (773) \\
\midrule
\dftavg{} & -- & -- & -- & -- & -- \\
\midrule
OXtal & 30 & 30 & 0.300 & 0.220 & -- \\
\name{}-M & 30 & 30 & 0.697 $\pm$ 0.039 & 0.241 $\pm$ 0.044 & 0.442 $\pm$ 0.010 \\
\name{}-M & 150 & 30 & 0.712 $\pm$ 0.035 & 0.260 $\pm$ 0.043 & 0.470 $\pm$ 0.010 \\
\name{}-L & 30 & 30 & 0.731 $\pm$ 0.040 & 0.287 $\pm$ 0.044 & 0.461 $\pm$ 0.010 \\
\name{}-L & 150 & 30 & 0.772 $\pm$ 0.039 & 0.346 $\pm$ 0.047 & 0.484 $\pm$ 0.010 \\
\name{}-H & 30 & 30 & 0.766 $\pm$ 0.034 & 0.378 $\pm$ 0.044 & 0.584 $\pm$ 0.010 \\
\name{}-H & 150 & 30 & \textbf{0.798 $\pm$ 0.028} & \textbf{0.479 $\pm$ 0.045} & \textbf{0.621 $\pm$ 0.009} \\
\midrule
\name{}-M & 400 & 200 & 0.879 $\pm$ 0.023 & 0.506 $\pm$ 0.028 & 0.646 $\pm$ 0.007 \\
\name{}-L & 400 & 200 & \textbf{0.919 $\pm$ 0.018} & 0.596 $\pm$ 0.039 & 0.669 $\pm$ 0.007 \\
\name{}-H & 400 & 200 & 0.891 $\pm$ 0.020 & \textbf{0.678 $\pm$ 0.032} & \textbf{0.771 $\pm$ 0.007} \\
\midrule
\name{}-M & 1000 & 1000 & \textbf{0.960} & 0.620 & 0.754 \\
\name{}-L & 1000 & 1000 & 0.940 & 0.760 & 0.763 \\
\name{}-H & 1000 & 1000 & \textbf{0.960} & \textbf{0.780} & \textbf{0.856} \\
\bottomrule
\end{tabular}

\vspace{\baselineskip}

\begin{tabular}{lrr c c c}
\toprule
 & & & CSP5 & CSP6 & CSP7 \\
Method & $n_s$ & $k$ & (6) & (5) & (8) \\
\midrule
\dftavg{} & -- & -- & 0.544 & 0.496 & 0.421 \\
\midrule
OXtal & 30 & 30 & 0.167 & 0.200 & 0.125 \\
\name{}-M & 30 & 30 & 0.554 $\pm$ 0.140 & 0.311 $\pm$ 0.134 & 0.210 $\pm$ 0.089 \\
\name{}-M & 150 & 30 & 0.616 $\pm$ 0.134 & 0.374 $\pm$ 0.153 & 0.225 $\pm$ 0.080 \\
\name{}-L & 30 & 30 & 0.681 $\pm$ 0.140 & 0.355 $\pm$ 0.148 & 0.245 $\pm$ 0.096 \\
\name{}-L & 150 & 30 & \textbf{0.789 $\pm$ 0.142} & 0.480 $\pm$ 0.155 & 0.263 $\pm$ 0.091 \\
\name{}-H & 30 & 30 & 0.705 $\pm$ 0.120 & 0.445 $\pm$ 0.149 & 0.280 $\pm$ 0.119 \\
\name{}-H & 150 & 30 & 0.785 $\pm$ 0.113 & \textbf{0.612 $\pm$ 0.152} & \textbf{0.393 $\pm$ 0.137} \\
\midrule
\name{}-M & 400 & 200 & 0.863 $\pm$ 0.099 & 0.657 $\pm$ 0.139 & 0.384 $\pm$ 0.066 \\
\name{}-L & 400 & 200 & \textbf{0.975 $\pm$ 0.061} & 0.729 $\pm$ 0.104 & 0.566 $\pm$ 0.128 \\
\name{}-H & 400 & 200 & 0.965 $\pm$ 0.069 & \textbf{0.766 $\pm$ 0.077} & \textbf{0.822 $\pm$ 0.126} \\
\midrule
\name{}-M & 1000 & 1000 & \textbf{1.000} & \textbf{0.800} & 0.500 \\
\name{}-L & 1000 & 1000 & \textbf{1.000} & \textbf{0.800} & 0.875 \\
\name{}-H & 1000 & 1000 & \textbf{1.000} & \textbf{0.800} & \textbf{1.000} \\
\bottomrule
\end{tabular}

\end{table}

\begin{table}[!htbp]
\centering
\addedcolor
\small
\caption{\added{Extended OXtal metrics on the OXtal benchmark subsets. When $k < n_s$, our methods select the top-$k$ by UMA energy.}}
\label{tab:oxtal-extended}
\begin{tabular}{llrrccccc}
\toprule
Subset & Method & $n_s$ & $k$ & Col$_S$ $\downarrow$ & Pac$_S$ $\uparrow$ & Pac$_C$ $\uparrow$ & Rec$_S$ $\uparrow$ & Rec$_C$ $\uparrow$ \\
\midrule
\multirow{7}{*}{Rigid} & OXtal & 30 & 30 & 0.011 & 0.873 & 1.000 & 0.737 & 0.960 \\
\cmidrule(l){2-9}
 & \name{}-M & 30 & 30 & 0.000 & 0.276 & 0.787 & 0.921 & 0.992 \\
 & \name{}-L & 30 & 30 & 0.000 & 0.292 & 0.819 & 0.923 & 1.000 \\
 & \name{}-H & 30 & 30 & 0.000 & 0.345 & 0.822 & 0.929 & 0.999 \\
\cmidrule(l){2-9}
 & \name{}-M & 150 & 30 & 0.000 & 0.303 & 0.802 & 0.918 & 0.986 \\
 & \name{}-L & 150 & 30 & 0.000 & 0.323 & 0.852 & 0.924 & 1.000 \\
 & \name{}-H & 150 & 30 & 0.000 & 0.387 & 0.833 & 0.930 & 0.997 \\
\midrule
\multirow{7}{*}{Flexible} & OXtal & 30 & 30 & 0.097 & 0.291 & 0.900 & 0.048 & 0.400 \\
\cmidrule(l){2-9}
 & \name{}-M & 30 & 30 & 0.000 & 0.034 & 0.347 & 0.273 & 0.828 \\
 & \name{}-L & 30 & 30 & 0.000 & 0.042 & 0.394 & 0.266 & 0.805 \\
 & \name{}-H & 30 & 30 & 0.000 & 0.064 & 0.458 & 0.339 & 0.869 \\
\cmidrule(l){2-9}
 & \name{}-M & 150 & 30 & 0.000 & 0.041 & 0.369 & 0.272 & 0.826 \\
 & \name{}-L & 150 & 30 & 0.000 & 0.056 & 0.441 & 0.278 & 0.804 \\
 & \name{}-H & 150 & 30 & 0.000 & 0.094 & 0.553 & 0.377 & 0.884 \\
\midrule
\multirow{7}{*}{CSP5} & OXtal & 30 & 30 & 0.006 & 0.667 & 0.833 & 0.572 & 0.833 \\
\cmidrule(l){2-9}
 & \name{}-M & 30 & 30 & 0.000 & 0.087 & 0.639 & 0.826 & 0.891 \\
 & \name{}-L & 30 & 30 & 0.000 & 0.103 & 0.749 & 0.816 & 0.938 \\
 & \name{}-H & 30 & 30 & 0.000 & 0.252 & 0.737 & 0.822 & 0.913 \\
\cmidrule(l){2-9}
 & \name{}-M & 150 & 30 & 0.000 & 0.097 & 0.671 & 0.828 & 0.903 \\
 & \name{}-L & 150 & 30 & 0.000 & 0.114 & 0.828 & 0.824 & 0.940 \\
 & \name{}-H & 150 & 30 & 0.000 & 0.286 & 0.807 & 0.830 & 0.883 \\
\midrule
\multirow{7}{*}{CSP6} & OXtal & 30 & 30 & 0.013 & 0.660 & 1.000 & 0.160 & 0.600 \\
\cmidrule(l){2-9}
 & \name{}-M & 30 & 30 & 0.000 & 0.200 & 0.402 & 0.544 & 0.868 \\
 & \name{}-L & 30 & 30 & 0.000 & 0.200 & 0.445 & 0.556 & 0.908 \\
 & \name{}-H & 30 & 30 & 0.000 & 0.211 & 0.521 & 0.572 & 0.917 \\
\cmidrule(l){2-9}
 & \name{}-M & 150 & 30 & 0.000 & 0.208 & 0.533 & 0.565 & 0.911 \\
 & \name{}-L & 150 & 30 & 0.000 & 0.210 & 0.512 & 0.569 & 0.868 \\
 & \name{}-H & 150 & 30 & 0.000 & 0.227 & 0.647 & 0.588 & 0.943 \\
\midrule
\multirow{7}{*}{CSP7} & OXtal & 30 & 30 & 0.021 & 0.483 & 0.875 & 0.129 & 0.375 \\
\cmidrule(l){2-9}
 & \name{}-M & 30 & 30 & 0.000 & 0.066 & 0.324 & 0.410 & 0.706 \\
 & \name{}-L & 30 & 30 & 0.000 & 0.071 & 0.366 & 0.394 & 0.728 \\
 & \name{}-H & 30 & 30 & 0.000 & 0.080 & 0.419 & 0.400 & 0.738 \\
\cmidrule(l){2-9}
 & \name{}-M & 150 & 30 & 0.000 & 0.071 & 0.307 & 0.419 & 0.712 \\
 & \name{}-L & 150 & 30 & 0.000 & 0.075 & 0.332 & 0.392 & 0.733 \\
 & \name{}-H & 150 & 30 & 0.000 & 0.111 & 0.521 & 0.436 & 0.780 \\
\bottomrule
\end{tabular}

\end{table}

\FloatBarrier

\added{\subsection{Polymorph coverage}
\label{app:polymorphs}}

\added{To evaluate whether \name{} recovers multiple polymorphs, we collect the $48$ refcode families from the OXtal test set that have multiple members. For each family, we draw $n_s = 150$ samples from \name{}-H and match them against all members. The coverage of a family is the fraction of members that are matched (in the Sol sense) by at least one sample, which we average over families uniformly or size-weighted.
\name{}-H recovers $61$--$75\%$ of the observed polymorphs (Table~\ref{tab:polymorph}) using only $150$ samples per family without energy ranking, whereas template generation approaches routinely generate thousands of samples before refinement and down-selection \cite{yang2025genarris}. Coverage is stronger on the rigid families than on the flexible and blind-test families, showing room for future work.
Samples are also diverse: across the $48$ families, the family-balanced mean within-family EMD PDD is $0.208$ for experimental polymorphs and $0.412$ for $1000$ \name{}-H samples per target. Table~\ref{tab:polymorph-clusters} reports the number of clusters among these samples at different EMD PDD thresholds.}

\begin{table}[h]
\centering
\addedcolor
\small
\caption{\added{Polymorph coverage of \name{}-H on the $48$ OXtal test refcode families with multiple members, using $n_s = 150$ samples per family without energy ranking. Coverage is averaged over families uniformly or size-weighted.}}
\label{tab:polymorph}
\begin{tabular}{lcccccc}
\toprule
 & & & \multicolumn{2}{c}{Family size} & \multicolumn{2}{c}{Coverage} \\
\cmidrule(lr){4-5} \cmidrule(l){6-7}
Group & Families & Polymorphs & Median & [Min, Max] & Uniform & Weighted \\
\midrule
Rigid    & 30 & 379 & 4    & [2, 74]  & 0.757 & 0.871 \\
Flexible & 10 & 181 & 12.5 & [3, 53]  & 0.376 & 0.564 \\
CSP5     &  1 &  13 & 13   & [13, 13] & 0.231 & 0.231 \\
CSP6     &  2 &   7 & 3.5  & [2, 5]   & 0.300 & 0.429 \\
CSP7     &  5 &  11 & 2    & [2, 3]   & 0.400 & 0.364 \\
\midrule
All      & 48 & 591 & 4.5  & [2, 74]  & 0.611 & 0.748 \\
\bottomrule
\end{tabular}

\vspace{1.5\baselineskip}
\caption{\added{Median number of clusters among $1000$ \name{}-H samples per OXtal test target at different EMD PDD thresholds. Cluster counts are threshold-dependent and should not be interpreted as numbers of physical polymorphs.}}
\label{tab:polymorph-clusters}
\begin{tabular}{lccccccc}
\toprule
EMD PDD threshold & 0.10 & 0.20 & 0.25 & 0.30 & 0.35 & 0.40 & 0.50 \\
\midrule
Median clusters & 1000 & 978 & 851 & 525 & 157 & 38 & 2 \\
\bottomrule
\end{tabular}

\end{table}

\FloatBarrier

\added{\subsection{Supplementary figures}}

\added{In the following pages, we provide additional supporting figures.}

\clearpage
\vspace*{\fill}
\begin{figure}[h]
\centering
\includegraphics[width=0.9\linewidth]{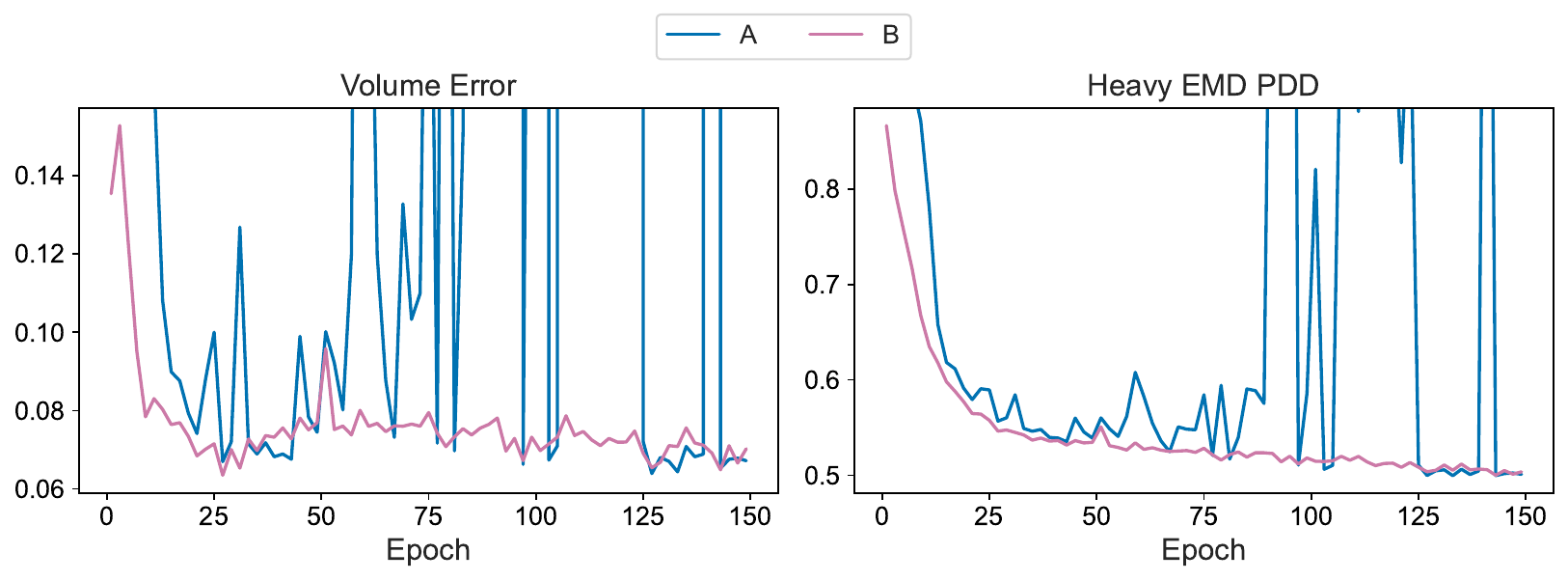}
\caption{Reconstruction metric curves on the validation set for models \textbf{A} and \textbf{B} (Table \ref{tab:ablation}). Results were pulled from preliminary training logs, where we only evaluated on a subset of 512 validation crystals, drew 3 samples per crystal, and did a sample-wise mean (not min.) aggregation.}
\label{fig:AB-loss-curve}
\end{figure}

\begin{figure}[h]
\centering
\includegraphics[width=0.5\linewidth]{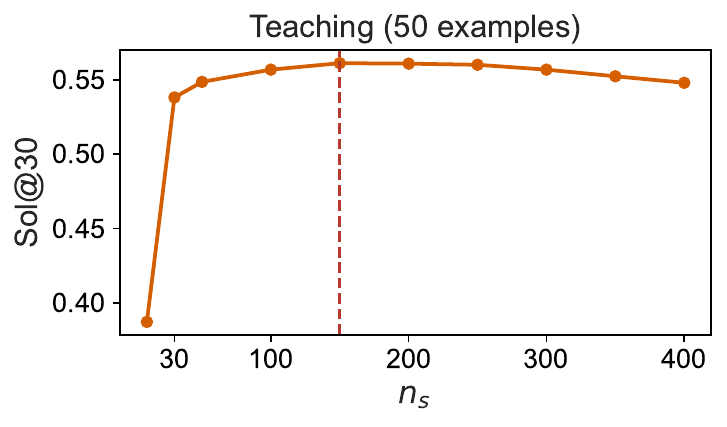}
\caption{Sol@30 as a function of $n_s$ on 50 random examples drawn from the CSD Teaching Subset. We find that $n_s = 150$ achieves the best solve rate for Sol@$30$. We sample 1000 times per example and report the mean over 5000 bootstrap resamples.}
\label{fig:estimate_ns}
\end{figure}
\vspace*{\fill}
\clearpage

\begin{figure}[p]
\centering
\begin{minipage}[t]{0.31\textwidth}
  \centering
  \includegraphics[width=\linewidth]{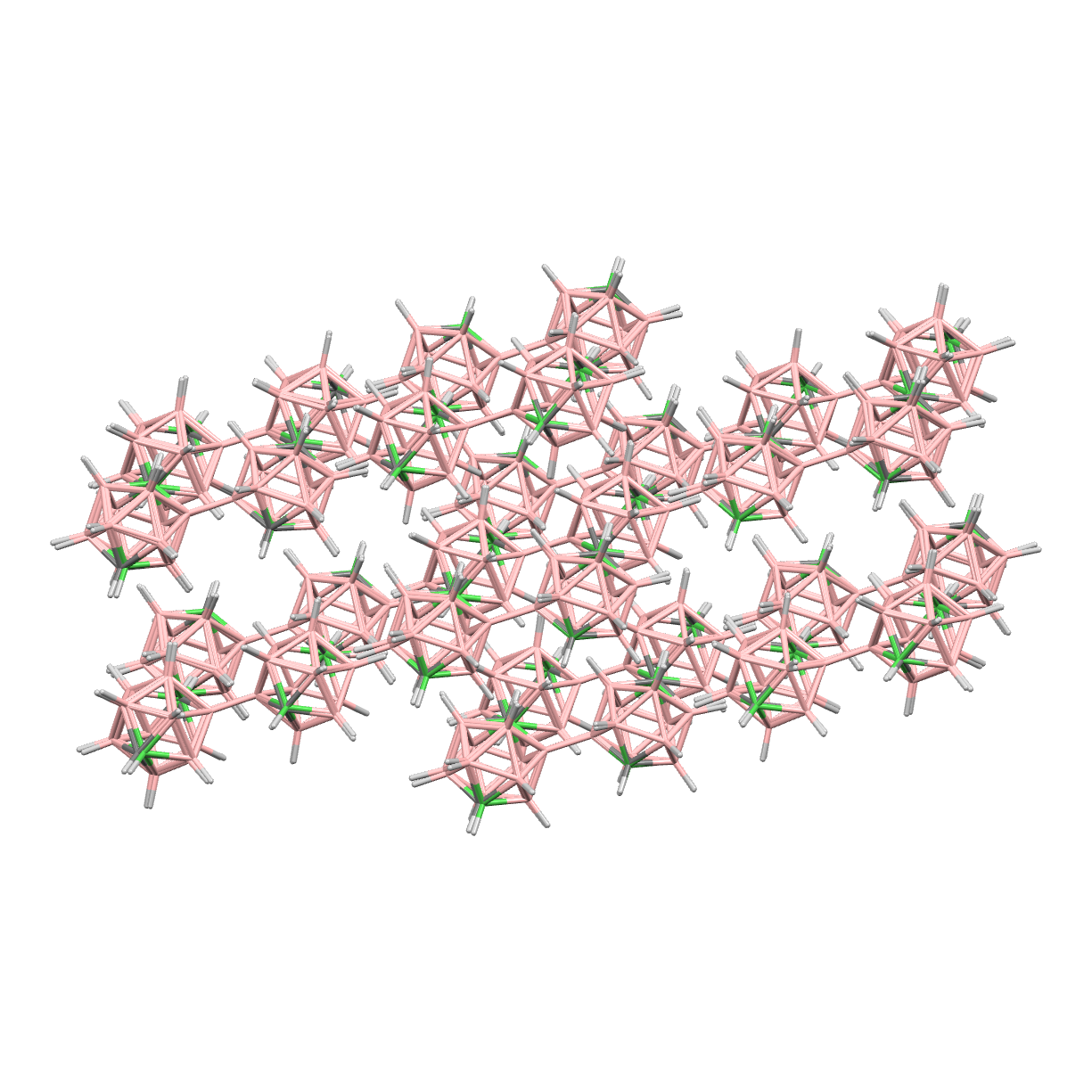}\\
  \vspace{1ex}
  \small (a) BCABOR10, $0.117$\,\AA
\end{minipage}\hfill
\begin{minipage}[t]{0.31\textwidth}
  \centering
  \includegraphics[width=\linewidth]{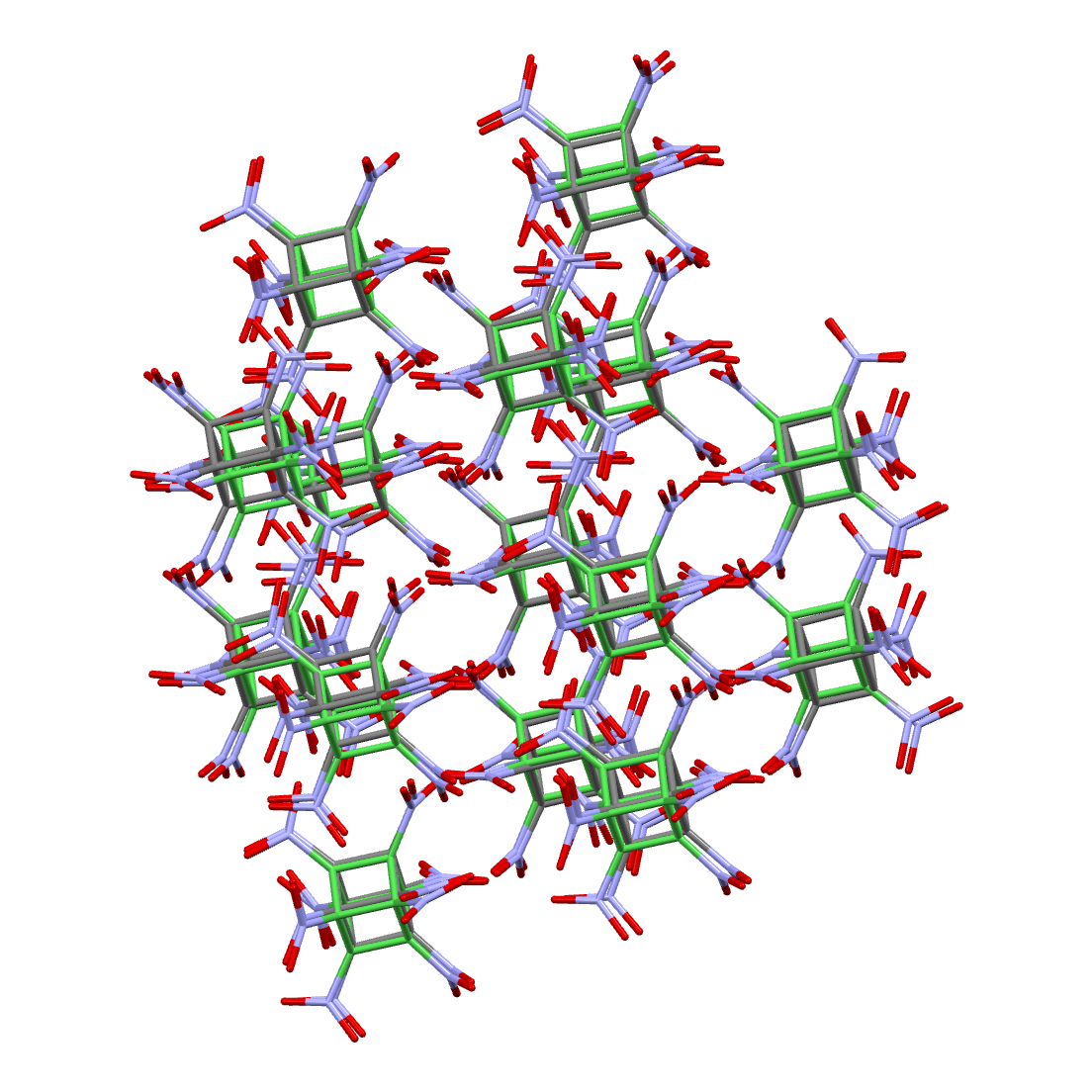}\\
  \vspace{1ex}
  \small (b) CUGDIR, $0.356$\,\AA
\end{minipage}\hfill
\begin{minipage}[t]{0.31\textwidth}
  \centering
  \includegraphics[width=\linewidth]{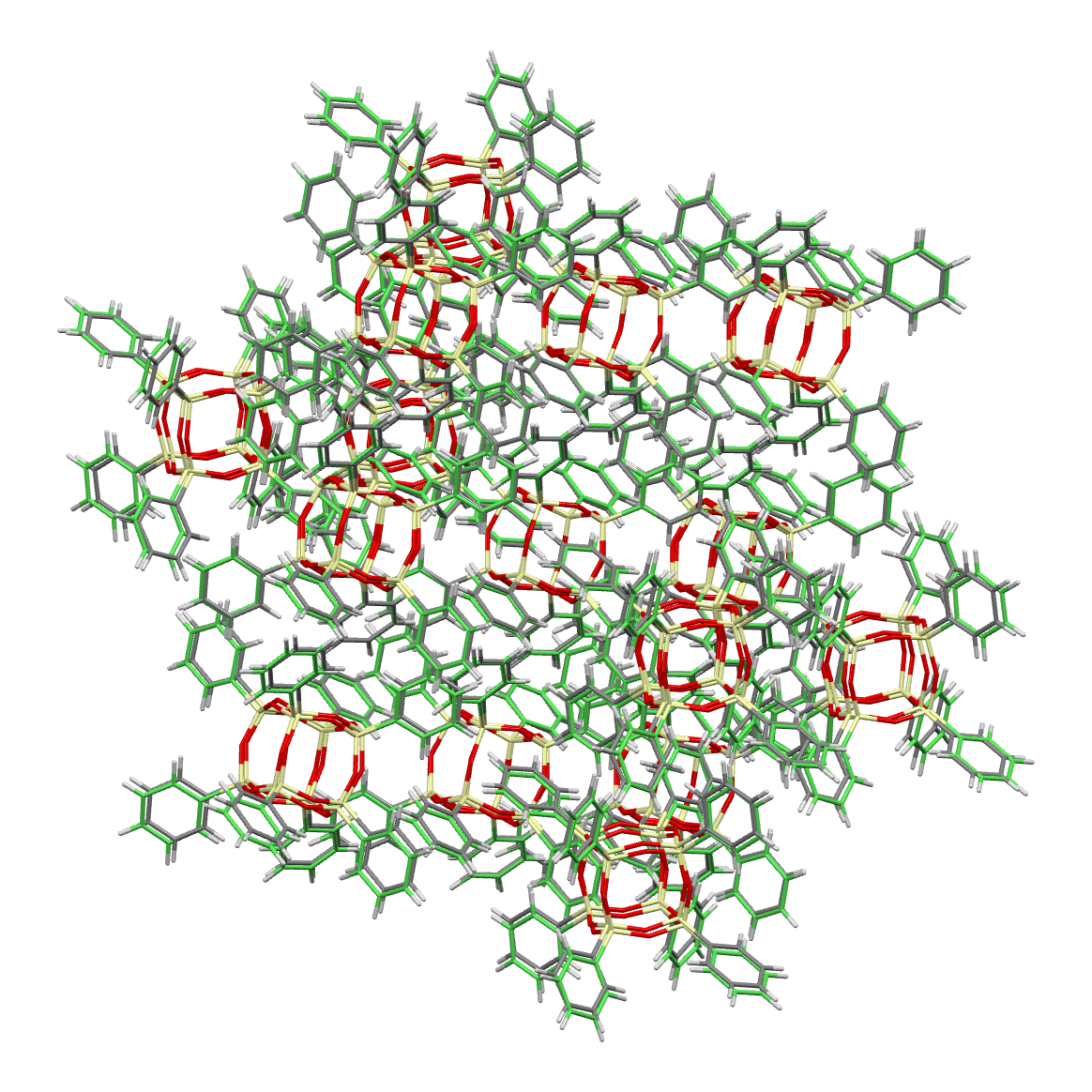}\\
  \vspace{1ex}
  \small (c) DORRAF, $0.252$\,\AA
\end{minipage}
\vspace{1ex}

\begin{minipage}[t]{0.31\textwidth}
  \centering
  \includegraphics[width=\linewidth]{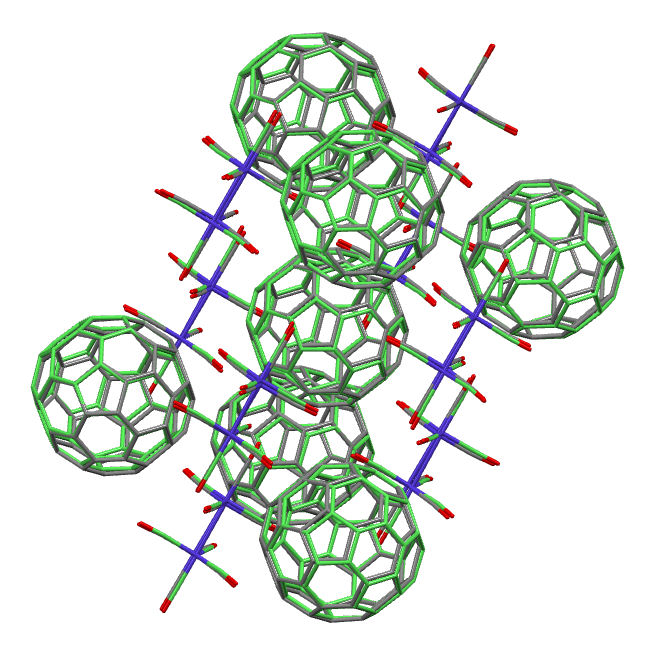}\\
  \vspace{1ex}
  \small (d) YUGWUT, $0.226$\,\AA
\end{minipage}\hfill
\begin{minipage}[t]{0.31\textwidth}
  \centering
  \includegraphics[width=\linewidth]{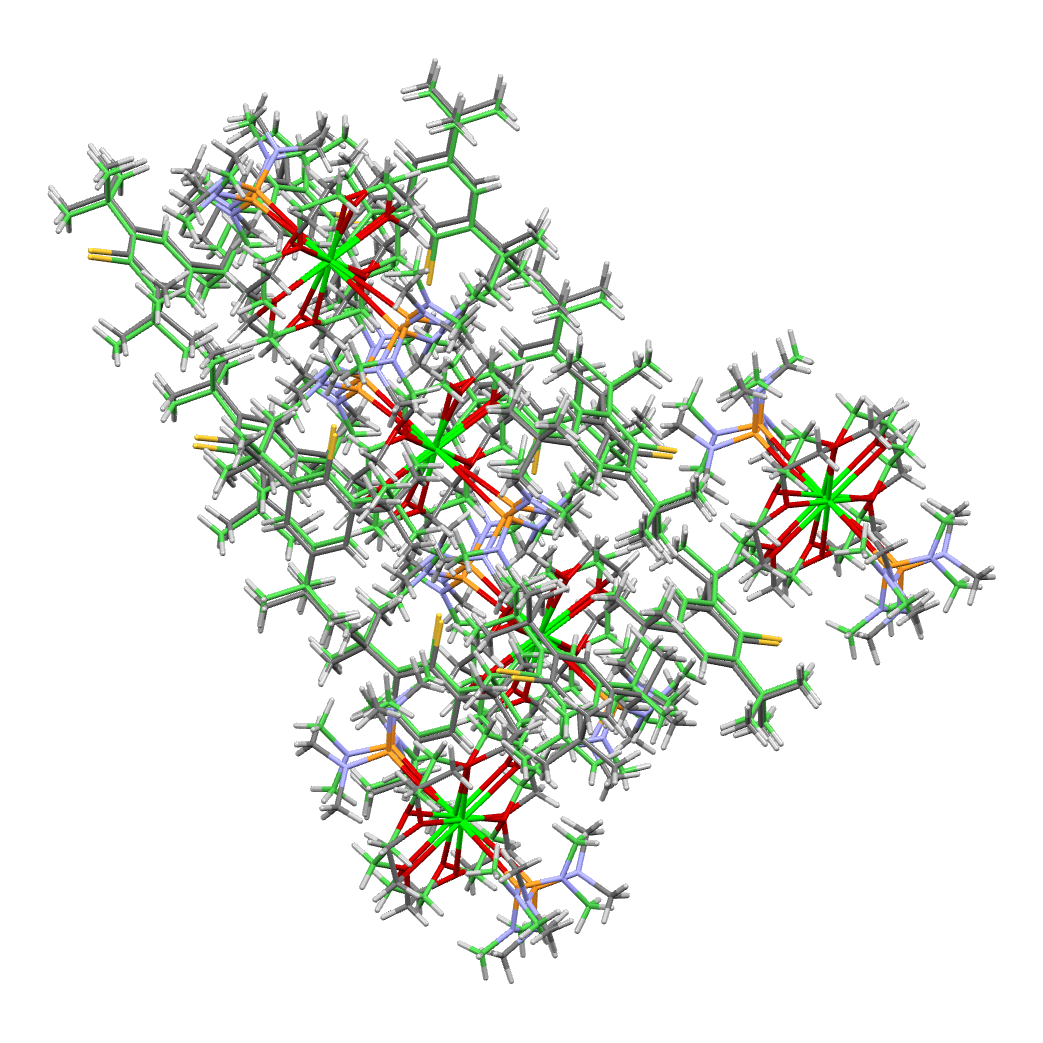}\\
  \vspace{1ex}
  \small (e) JIYDEA, $0.575$\,\AA
\end{minipage}\hfill
\begin{minipage}[t]{0.31\textwidth}
  \centering
  \includegraphics[width=\linewidth]{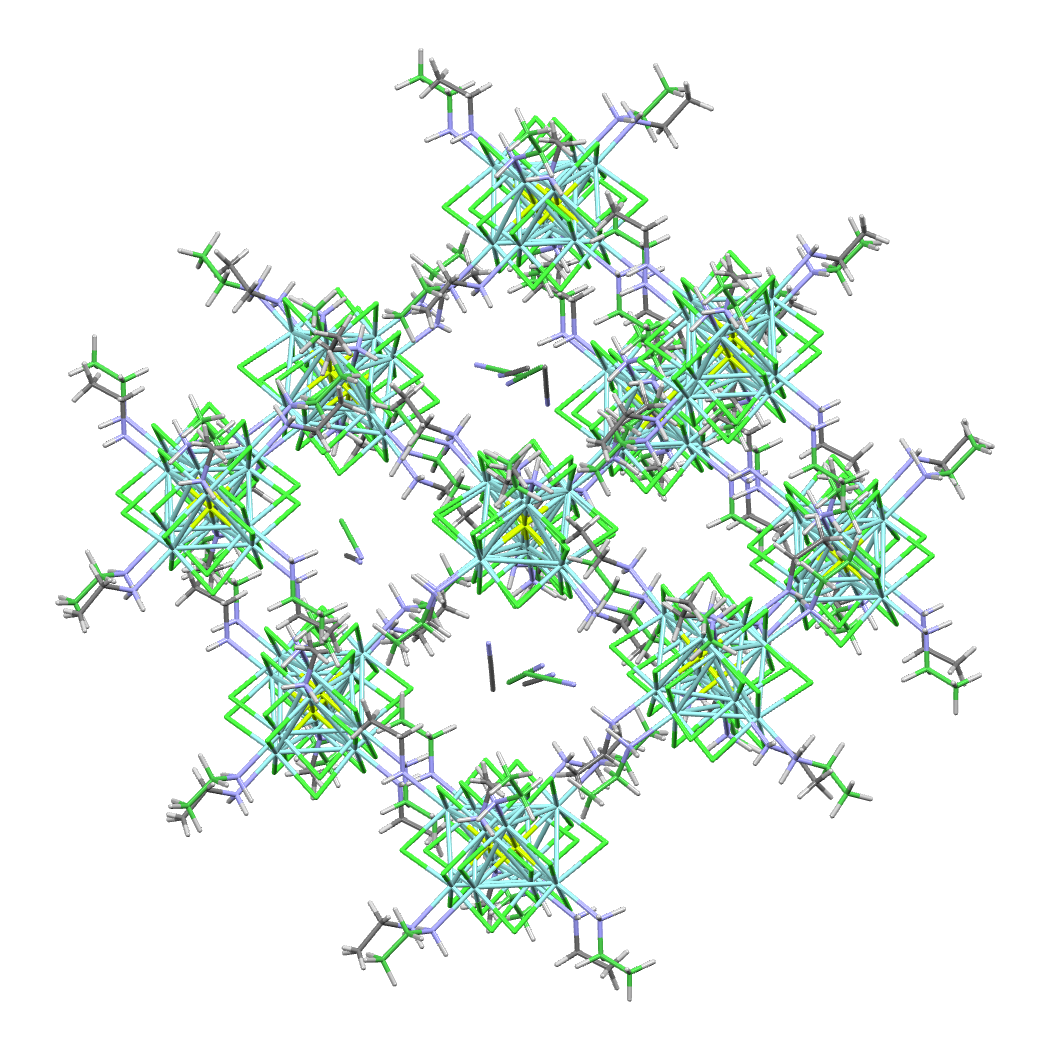}\\
  \vspace{1ex}
  \small (f) SIMLIJ, $1.082$\,\AA
\end{minipage}
\vspace{1ex}

\begin{minipage}[t]{0.31\textwidth}
  \centering
  \includegraphics[width=\linewidth]{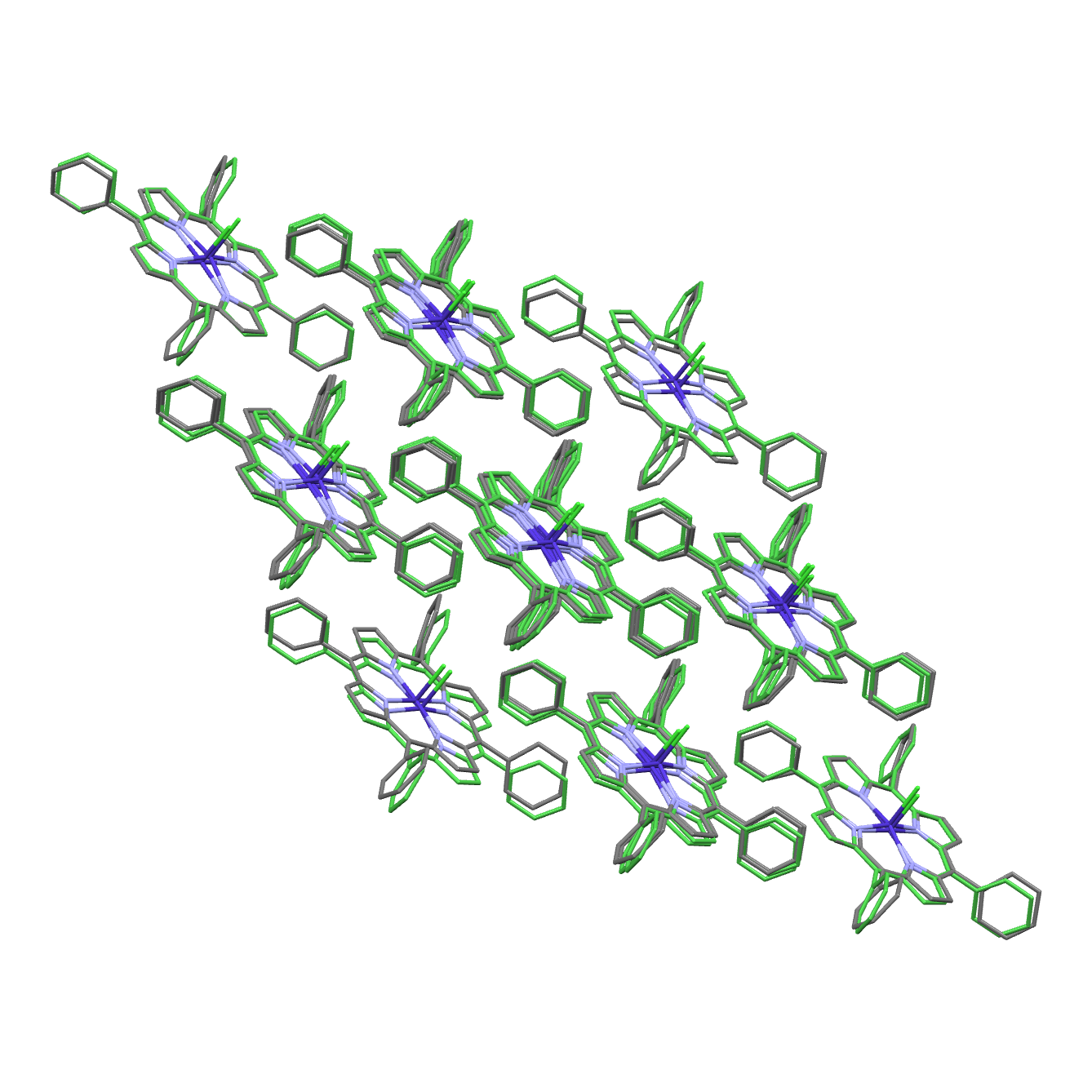}\\
  \vspace{1ex}
  \small (g) CTPOCO, $0.445$\,\AA
\end{minipage}\hfill
\begin{minipage}[t]{0.31\textwidth}
  \centering
  \includegraphics[width=\linewidth]{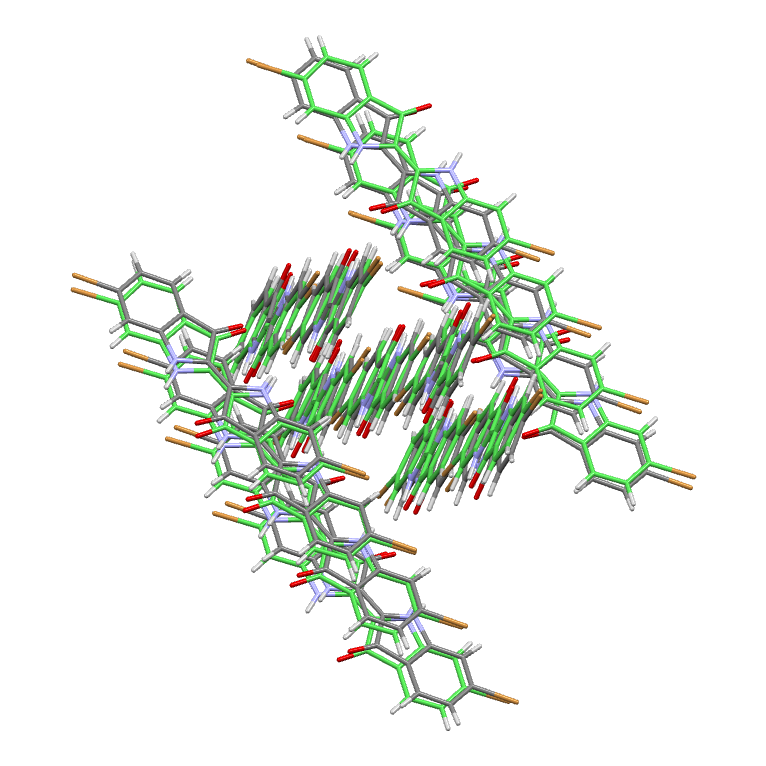}\\
  \vspace{1ex}
  \small (h) DBRING01, $0.442$\,\AA
\end{minipage}\hfill
\begin{minipage}[t]{0.31\textwidth}
  \centering
  \includegraphics[width=\linewidth]{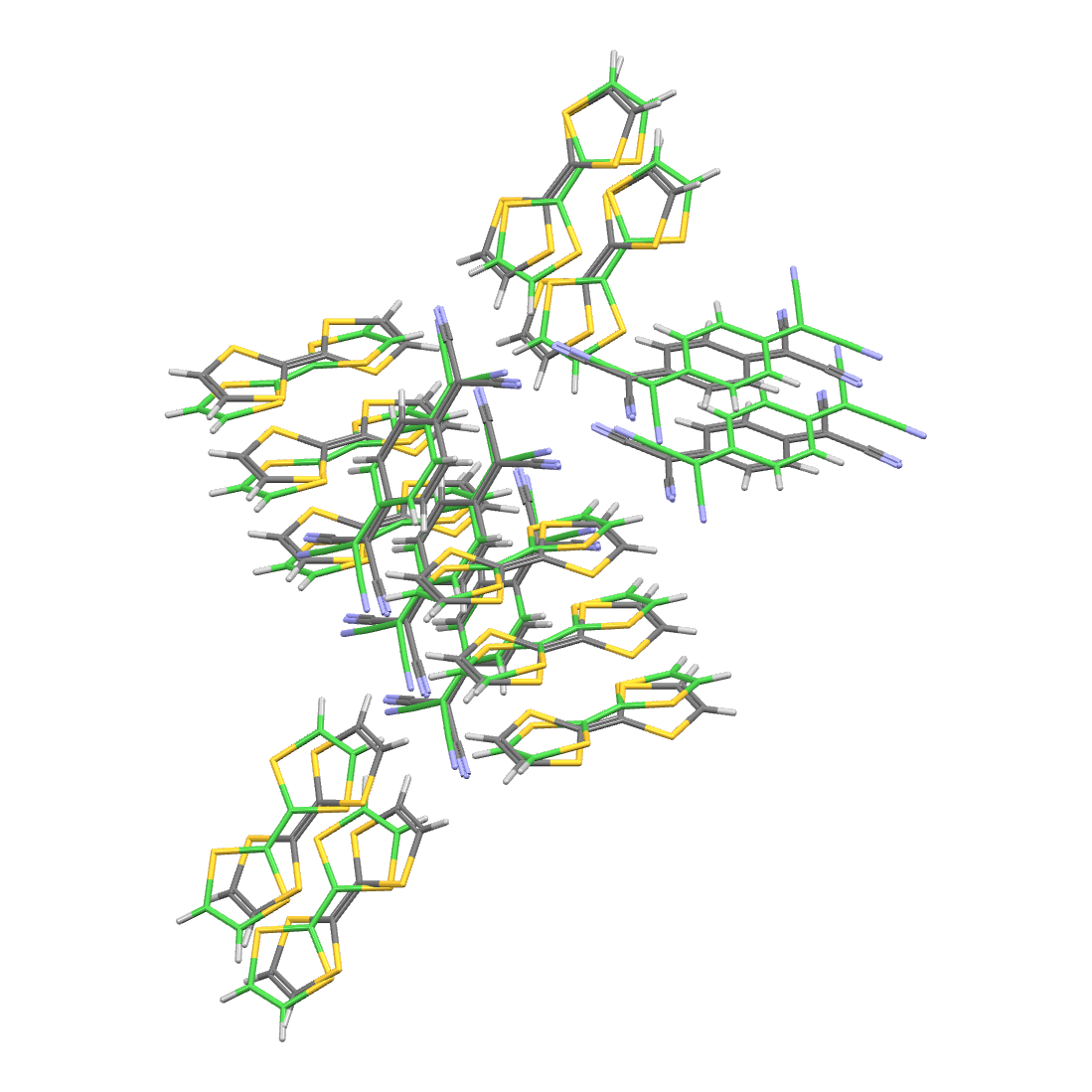}\\
  \vspace{1ex}
  \small (i) TTFTCQ, $0.739$\,\AA
\end{minipage}
\vspace{1ex}

\caption{Predicted crystal structures from the CSD Teaching Subset, all achieving full 15/15 molecule matches in COMPACK with the reported RMSD$_{15}$ values (\AA).}
\label{fig:compack}
\end{figure}

\end{document}